\documentclass[12pt,a4paper]{article}

\usepackage{graphicx}

\usepackage{latexsym,rotating,calc,textcase,booktabs,color,endnotes}
\usepackage{amsfonts,amssymb,amsbsy,amsmath,amsthm}
\usepackage{bm}
\usepackage{multirow}
\usepackage{subcaption}
\usepackage{algorithm}
\usepackage[noEnd]{algpseudocodex}
\algrenewcommand\alglinenumber[1]{\footnotesize #1}
\algrenewcommand\algorithmicdo{}
\usepackage{enumitem}
\usepackage{placeins}
\usepackage[numbers,square,sort&compress]{natbib}
\usepackage[colorlinks=true,linkcolor=blue,citecolor=blue,urlcolor=blue]{hyperref}

\newenvironment{keywords}{\small\textbf{Keywords:}\;}{}

\newtheorem{theorem}{Theorem}[section]
\newtheorem{lemma}[theorem]{Lemma}
\newtheorem{corollary}[theorem]{Corollary}
\newtheorem{remark}[theorem]{Remark}

\newcommand{\affilnum}[1]{}

\title{FMT\textsuperscript{X}: Lazy Wavefront Search for Dynamic Replanning}
\author{Soheil Espahbodi Nia\thanks{Email: soheil.e.nia@gmail.com}}
\date{}

\begin{document}
	\maketitle
	\markboth{Espahbodi Nia}{FMT\textsuperscript{X}: Lazy Wavefront Search for Dynamic Replanning}
	\begin{abstract}
		FMT$^{*}$ plans efficiently in static worlds by expanding a cost-ordered
		wavefront and collision-checking lazily, but its single-pass unvisited rule
		cannot revise paths when obstacles change.
		We present FMT\textsuperscript{X}, an anytime, asymptotically optimal
		generalization of that wavefront for dynamic replanning.
		A cost-improvement test replaces the unvisited set, allowing a node to be
		revisited for best-parent selection whenever a lower-cost potential connection is found.
		This induces implicit rewiring within the wavefront while preserving lazy
		collision checking.
		FMT\textsuperscript{X} retains the online densification of RRT$^{*}$ but
		defers its eager neighborhood rewiring to the cost-ordered wavefront, so a
		node is revisited only when the expansion reaches it.
		Obstacle updates orphan the affected subtree and reseed the wavefront.
		A direct cost push from each parent to its children propagates cost improvements
		through validated tree edges, thereby preserving descendant cost consistency,
		a property not guaranteed by implicit rewiring alone.
		We compare a fixed-graph batch variant, Dynamic FMT$^{*}$ (D-FMT$^{*}$), with
		D$^{*}$~Lite on identical PRM$^{*}$ graphs, and	FMT\textsuperscript{X} with eager RRT\textsuperscript{X} and path-centric
		LLPT$^{*}$ in geometric and kinodynamic scenes, including partial
		observability.
		D-FMT$^{*}$ stays close on path quality at far fewer collision checks.
		FMT\textsuperscript{X} lies between LLPT$^{*}$ and RRT\textsuperscript{X} on
		repair effort while tracking the reliability and trajectory quality of eager
		RRT\textsuperscript{X}.
		The study shows that collision-checking policy affects median repair latency,
		repair-time tails, and executed trajectory quality.
	\end{abstract}


\section{Introduction}
\label{sec:introduction}

Motion planning among changing obstacles requires fast adaptation without
restarting from scratch on every update. Static precomputation goes stale,
while full replanning is often too costly for real time.
Sampling-based planners such as RRT$^{*}$~\cite{karaman2011sampling} and
FMT$^{*}$~\cite{janson2015fast} provide asymptotically optimal solutions in
static worlds.
FMT$^{*}$, in particular, performs well in challenging geometries by running a
lazy dynamic-programming recursion that defers expensive collision checks.
That efficiency, however, comes from a rigid one-pass construction that
cannot revise paths when the environment changes.

Local repair of a search structure, rather than restart, was developed in
graph search by LPA$^{*}$ and D$^{*}$~Lite~\cite{Koenig2004LifelongPA,Koenig2005FastRF}
and adapted to sampling-based planning in RRT\textsuperscript{X}~\cite{otte2016rrtx},
which applies selective rewiring inside the RRT$^{*}$ framework.
Architecturally, RRT$^{*}$ grows incrementally by sequential sampling, whereas
FMT$^{*}$ propagates a wavefront through a pre-sampled batch using lazy
dynamic programming.
Directly porting RRT\textsuperscript{X}'s recursive rewiring cascades to
FMT$^{*}$ would conflict with that ordered propagation and weaken the lazy
evaluation guarantees that drive FMT$^{*}$'s efficiency.

\begin{figure*}[!t]
	\centering
	
	\begin{subfigure}[b]{0.33\textwidth}
		\includegraphics[width=\textwidth]{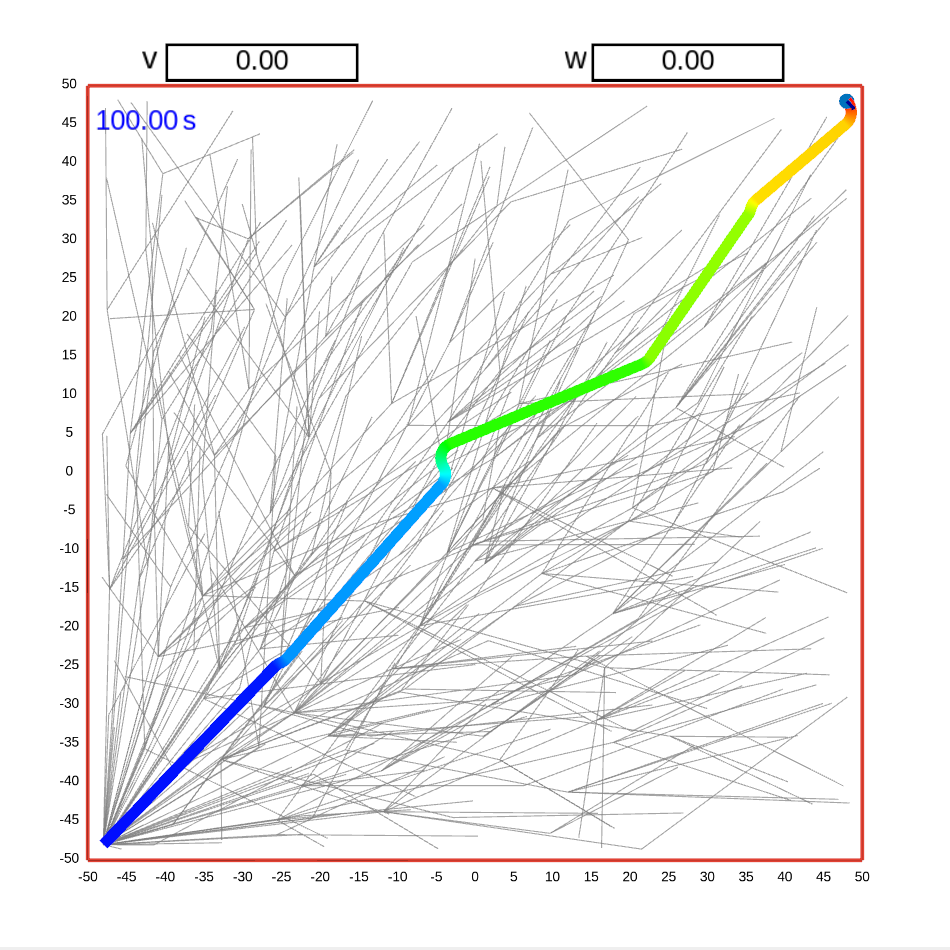}
	\end{subfigure}\hfill
	\begin{subfigure}[b]{0.33\textwidth}
		\includegraphics[width=\textwidth]{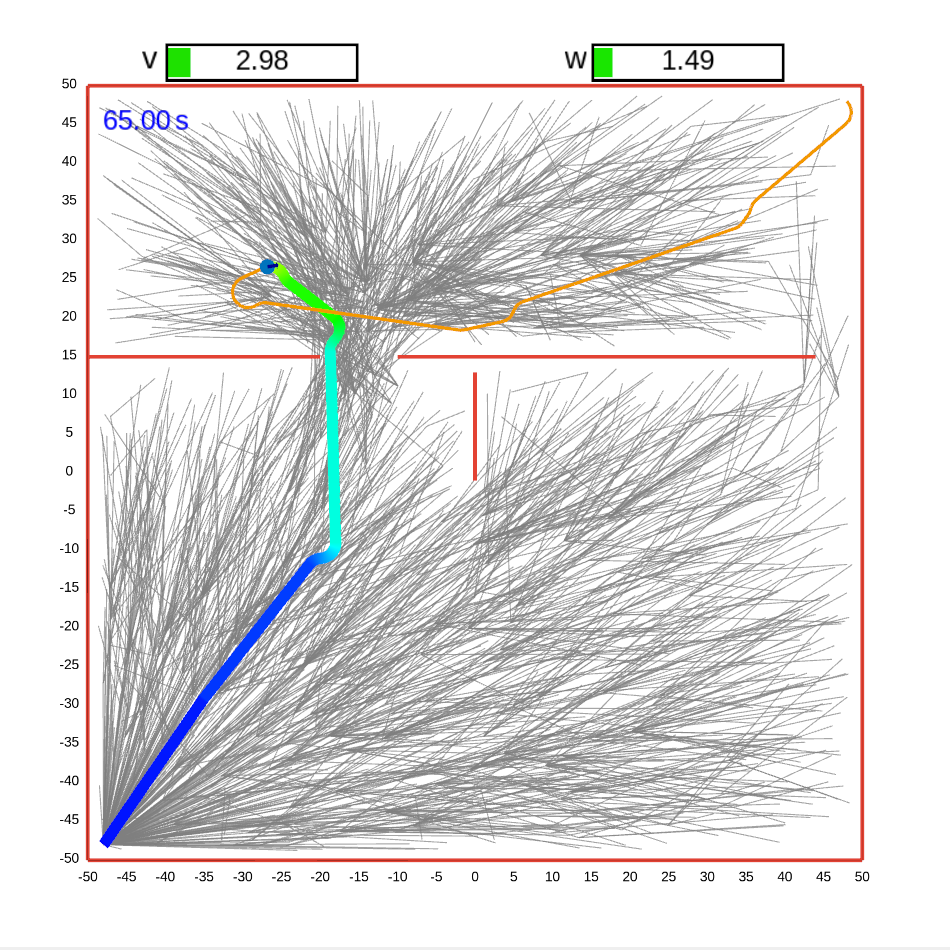}
	\end{subfigure}\hfill
	\begin{subfigure}[b]{0.33\textwidth}
		\includegraphics[width=\textwidth]{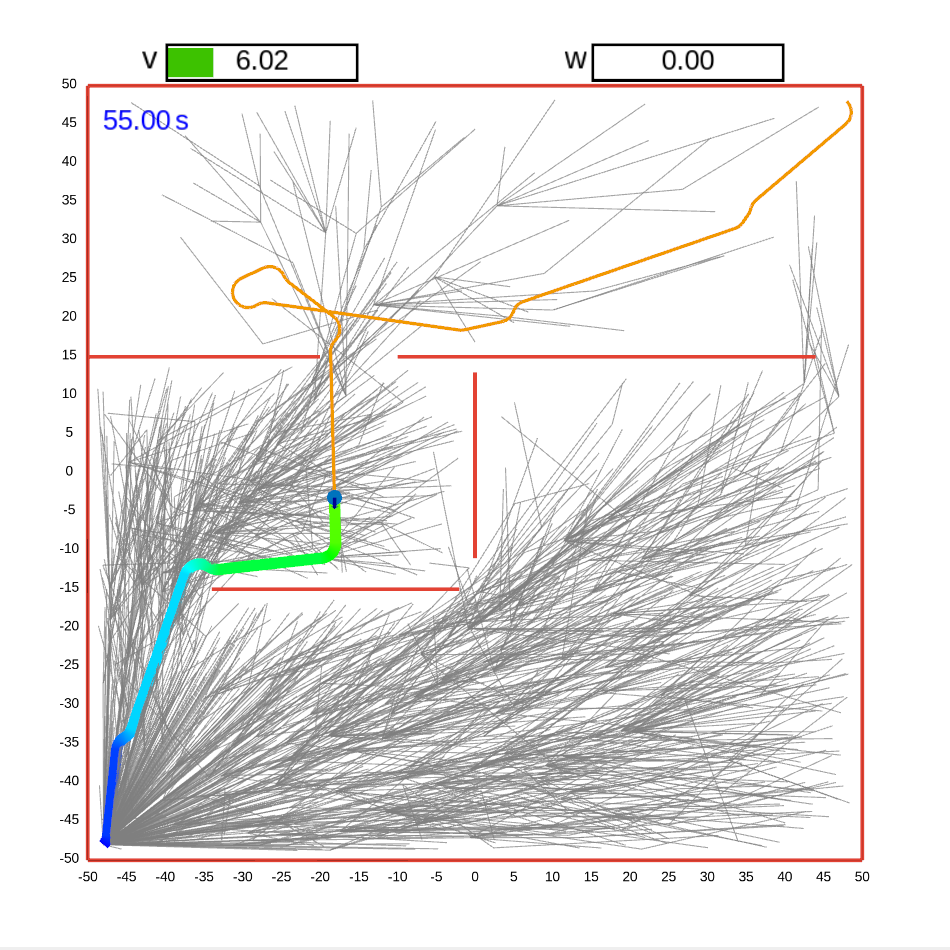}
	\end{subfigure}
	\\[4pt] 
	
	\begin{subfigure}[b]{0.33\textwidth}
		\includegraphics[width=\textwidth]{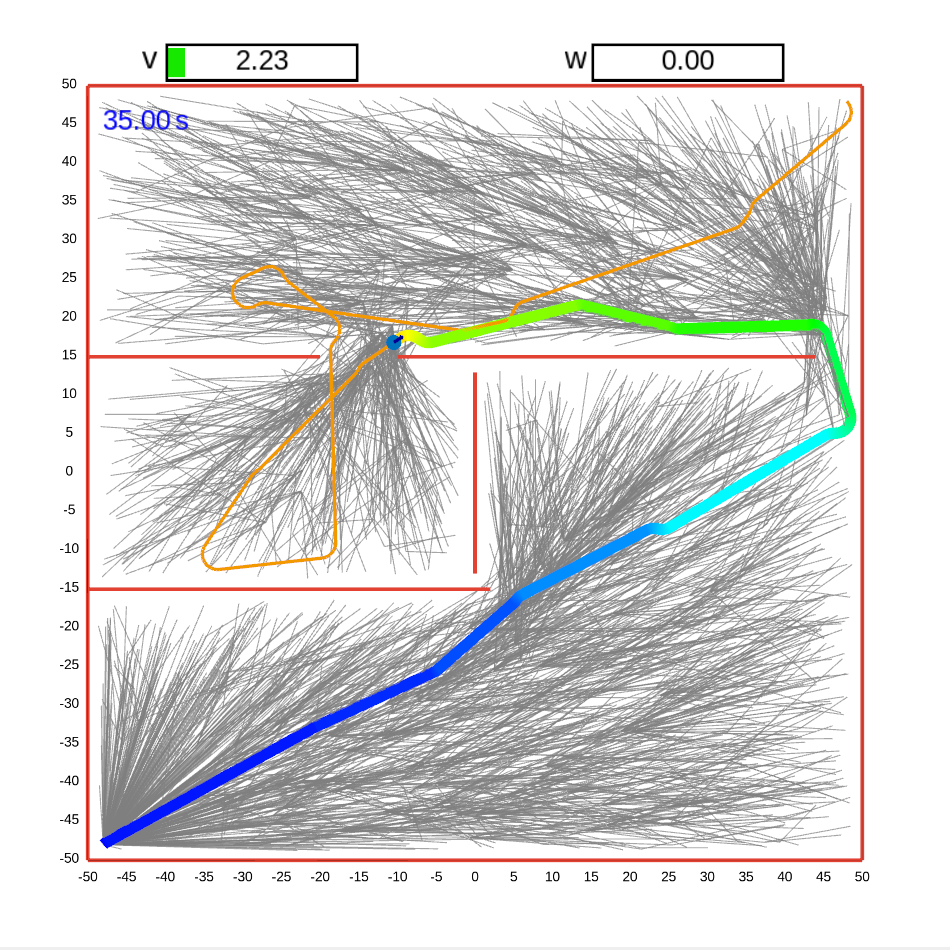}
	\end{subfigure}\hfill
	\begin{subfigure}[b]{0.33\textwidth}
		\includegraphics[width=\textwidth]{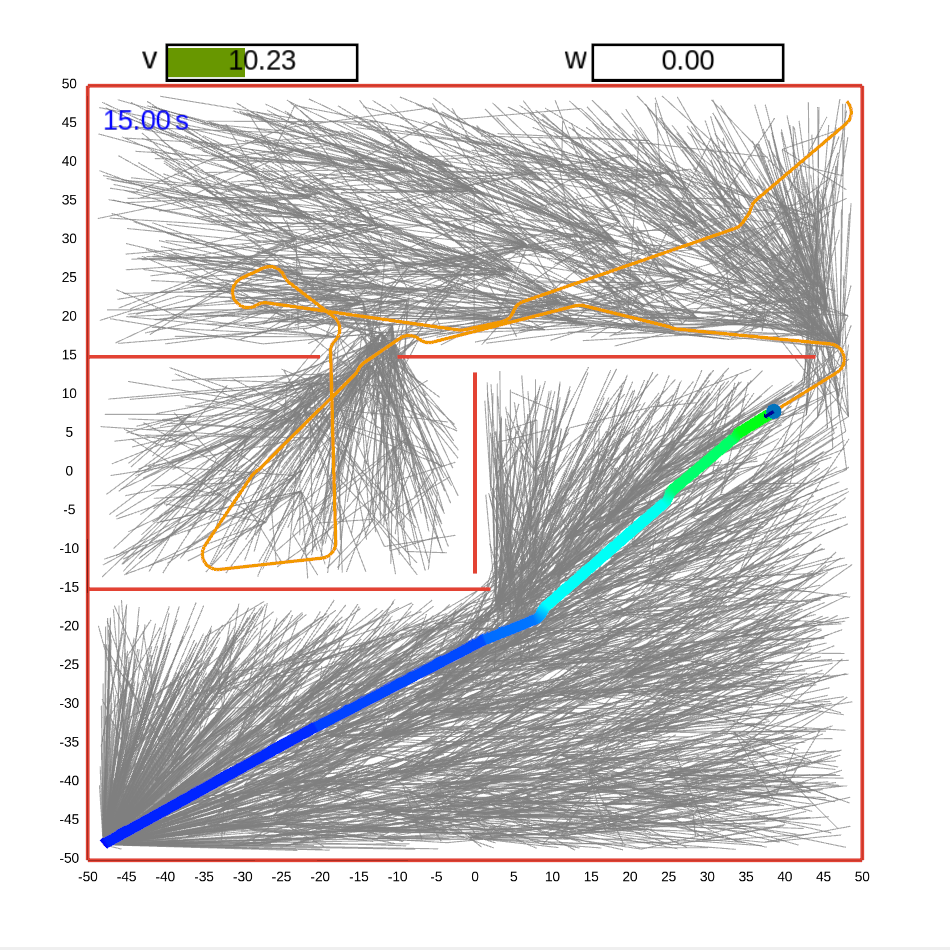}
	\end{subfigure}\hfill
	\begin{subfigure}[b]{0.33\textwidth}
		\includegraphics[width=\textwidth]{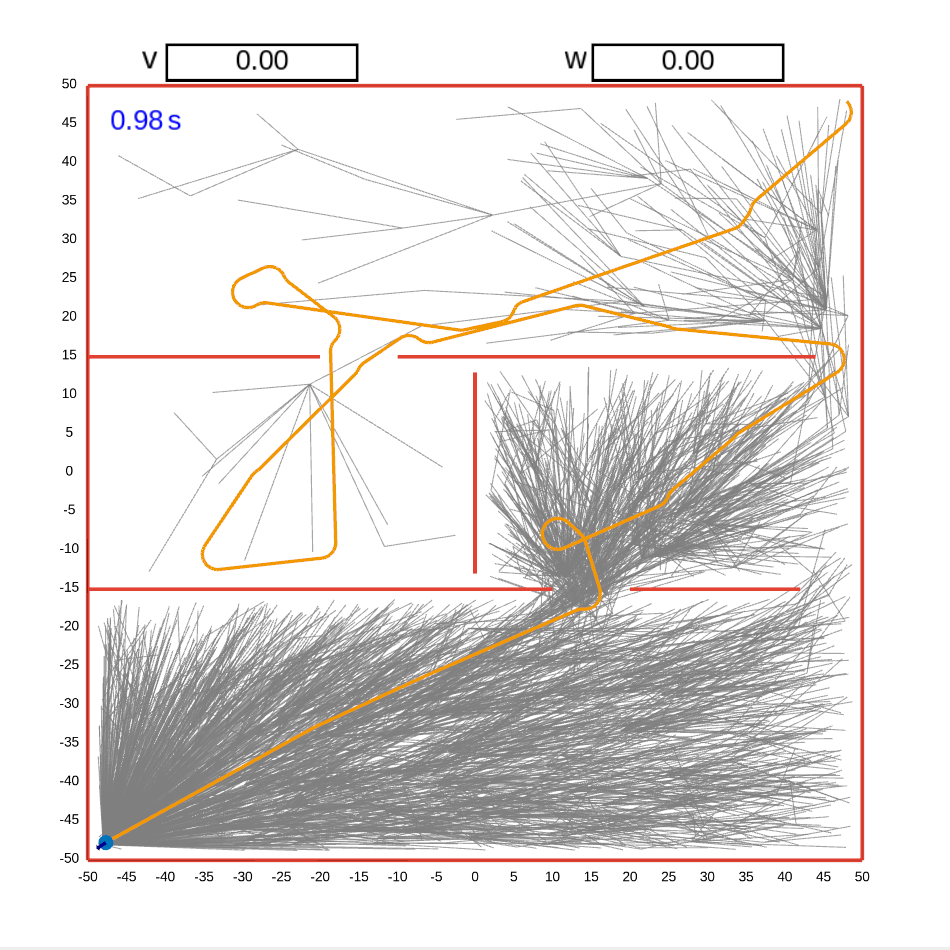}
	\end{subfigure}
	\\[4pt] 
	
	\caption{A sequential visualization of the FMT\textsuperscript{X} replanning process for the Dubins model in a partially observable environment with a limited sensor range. Edges are shown as straight lines between nodes for visualization only.}
	\label{fig:dubins_evolution}
\end{figure*}

We present FMT\textsuperscript{X}, a principled generalization of the Fast
Marching Tree algorithm for dynamic and anytime replanning.
Our core insight is that the standard unvisited constraint of FMT$^{*}$ is a
special case of a more fundamental cost-consistency condition governing
optimal path expansion.
Unlike hybrid approaches~\cite{xu2020informed} that interleave lazy and
non-lazy operations, FMT\textsuperscript{X} replaces this rigid boolean check
with a cost inequality that serves as a provable, event-driven indicator of
potential suboptimality.
Crucially, this inequality does not force immediate rewiring to the triggering
node. Instead, it certifies that the node's current cost
may be outdated and justifies a localized Bellman re-evaluation among all
valid open neighbors.
This introduces implicit rewiring while retaining the lazy collision-checking
strategy of the original algorithm.
This places FMT\textsuperscript{X} at a specific point in the design space. It
retains the online densification of RRT$^{*}$, but defers its eager
neighborhood rewiring to a cost-ordered lazy wavefront, so a node is revisited
only when the expansion reaches it. Whether this deferral
preserves asymptotic optimality is the central theoretical question raised by
the design, and Section~\ref{sec:runtime_analysis} establishes that
optimality is retained.
Lemma~\ref{lem:cost-equivalence} establishes that, in the standard
from-scratch forward construction on a fixed sample set in an obstacle-free
environment, every connected node has its exact disk-graph cost.
Consequently, the strict cost-improvement test cannot trigger a revisit after
connection, so the generalized rule introduces no additional repair work in
this baseline case and retains the one-pass behavior of FMT$^{*}$.

However, relying solely on this implicit rewiring exposes a structural
vulnerability tied to the lazy collision-checking nature of FMT$^{*}$.
We call this situation cost-propagation failure. It occurs when a parent-cost
improvement does not reach its descendants during lazy repair, leaving their
subtree labels outdated.
In rare but critical cases during dynamic replanning, lazy evaluation can
sever the cost-update chain in exactly this way.
To overcome this without sacrificing computational efficiency,
FMT\textsuperscript{X} introduces a direct parent-to-child propagation step.
By eagerly pushing valid cost improvements down the established ancestry
before neighborhood evaluation begins, FMT\textsuperscript{X} keeps Bellman
updates alive on the search tree and prevents cost-propagation failure from
stranding descendants.

To support unbounded continuous operation, FMT\textsuperscript{X} further
integrates $\epsilon$-consistency and neighbor-culling strategies.
These mechanisms ensure that as the state-space graph grows infinitely, the
expected neighborhood size and the depth of cost-propagation cascades remain
computationally bounded, preserving the algorithm's $\mathcal{O}(n\log n)$
complexity.
We evaluate FMT\textsuperscript{X} in two complementary settings.
First, to isolate the efficacy of the graph-repair mechanism, we compare a
fixed-graph batch variant (D-FMT$^{*}$) against
D$^{*}$~Lite on a PRM$^{*}$ graph with identical sample set and topology.
Second, we benchmark FMT\textsuperscript{X} against
RRT\textsuperscript{X} and path-centric LLPT$^{*}$ in environments with
unpredictable moving obstacles, including partial observability.
Figure~\ref{fig:dubins_evolution} illustrates the method in operation on a
Dubins vehicle under limited sensing, where obstacles become visible only within
a fixed range, so that each discovery forces repair against geometry the planner
has not previously seen.
Empirical evaluation shows that D-FMT$^{*}$ stays close on path quality at far
fewer collision checks and that FMT\textsuperscript{X} lies between LLPT$^{*}$ and
RRT\textsuperscript{X} on repair effort while tracking the reliability and
trajectory quality of eager RRT\textsuperscript{X}.

\section{Related Work}
Motion planning balances trade-offs between optimality, completeness, and computational efficiency through two foundational paradigms. Search-based planning discretizes the environment into a graph to find optimal paths, while sampling-based planning builds probabilistic structures to navigate complex, high-dimensional problems without explicit discretization~\cite{Orthey2023SamplingBasedMP}. This section reviews algorithms from both domains.

\subsection{Search-Based Planners}

\subsubsection{Classical Planning Algorithms}
Graph-based planners discretize the configuration space, typically via uniform grids. Uninformed methods like Dijkstra's algorithm~\cite{dijkstra1959note} ensure optimality by exploring in order of path cost. Informed search techniques like A*~\cite{hart1968formal} accelerate planning by combining cost-to-come with an admissible heuristic. These approaches are powerful but primarily suited to static, known environments.

\subsubsection{Replanning and Dynamic Search Algorithms}
Dynamic environments require planners that efficiently repair solutions rather than planning from scratch. Algorithms such as D$^{*}$~\cite{Stentz1994TheDA} and D$^{*}$~Lite~\cite{Koenig2002Dlite} propagate cost changes backward from the goal to repair the search tree. Similarly, Lifelong Planning A* (LPA*)~\cite{Koenig2004LifelongPA} and Adaptive A*~\cite{koenig2006real} optimize repeated queries by updating heuristics and preserving valid search sections.

These algorithms established influential concepts such as lazy edge evaluation and local repair that serve as conceptual bridges to modern sampling-based replanning. However, graph-based planners scale poorly in high-DOF continuous environments, motivating sampling-based approaches that extend these repair principles to continuous spaces.

\subsection{Sampling-Based Planners}

\subsubsection{Static Sampling-Based Planners}
The Probabilistic Roadmap (PRM)~\cite{kavraki1996probabilistic} and Rapidly-Exploring Random Tree (RRT)~\cite{LaValle1998RapidlyexploringRT} established probabilistic completeness for static environments. PRM constructs a multi-query graph capturing free-space connectivity. RRT builds a single-query tree by expanding towards random samples, creating a bias towards unexplored regions. Asymptotically optimal variants PRM$^{*}$ and RRT$^{*}$~\cite{karaman2011sampling} introduced incremental rewiring to guarantee path quality. RRT\textsuperscript{\#}~\cite{Arslan2013UseOR, Arslan2015DynamicPG} enhances this with vertex consistency to accelerate convergence, while Fast Marching Tree (FMT$^{*}$)~\cite{janson2013fmt, janson2015fast} performs lazy batch wavefront updates to converge faster than incremental rewiring.

To reduce the computational bottleneck of expensive collision checks, algorithms like Lazy PRM~\cite{Bohlin2000PathPU} pioneered deferred edge evaluation by building roadmaps that assume edges are collision-free, verifying them only when an optimistic shortest path is found. This path-centric laziness was later extended to asymptotically optimal planners in Lazy PRM$^{*}$~\cite{Hauser2015LazyCC}. The theoretical foundations of this paradigm were subsequently formalized by the LazySP class~\cite{Dellin2016TowardLO}, which Haghtalab et al.~\cite{Haghtalab2018ThePV} proved asymptotically optimal with $\mathcal{O}(n/p)$ query complexity. The Generalized Lazy Search (GLS) framework~\cite{Mandalika2019GeneralizedLS} further refined these principles by introducing event-based toggles to optimally balance search effort against edge evaluation. Additionally, bidirectional planners, such as RRT-Connect~\cite{Kuffner2000RRTconnectAE}, the asymptotically optimal RRT$^{*}$-Connect~\cite{Klemm2015RRTConnectFA} and Bidirectional FMT$^{*}$~\cite{Starek2015AnAS}, accelerate search by expanding trees from both start and goal.

Informed RRT$^{*}$~\cite{Gammell2014InformedRO, Gammell2017InformedSF} improves convergence by restricting sampling to a hyperellipsoid containing the optimal solution. Batch Informed Trees (BIT$^{*}$)~\cite{Gammell2017BatchIT, Gammell2014BatchIT} combine informed sampling with the batch processing of FMT$^{*}$. While effective, these informed methods (including ABIT$^{*}$~\cite{Strub2020AdvancedB}, AIT$^{*}$~\cite{Strub2020AdaptivelyIT}, and EIT$^{*}$~\cite{Strub2021AdaptivelyIT}) prune samples globally based on fixed obstacles, making them ill-suited for dynamic replanning where valid regions shift unpredictably.

\subsubsection{Kinodynamic Sampling-Based Planners}
Kinodynamic planning addresses differential constraints (e.g., velocity limits)~\cite{Gammell2020AsymptoticallyOS}. Forward-propagation methods, such as Kinodynamic RRT~\cite{lavalle2001randomized} and Stable Sparse RRT (SST)~\cite{li2016asymptotically}, explore by integrating dynamics forward, avoiding complex boundary value problems (BVPs). In contrast, steering-based methods use BVP solvers to connect states exactly, enabling asymptotically optimal algorithms like Kinodynamic RRT$^{*}$~\cite{Karaman2010OptimalKM, Webb2012KinodynamicRO} and Kinodynamic FMT$^{*}$~\cite{Schmerling2014OptimalSM}.

\subsubsection{Dynamic Sampling-Based Replanning}
Dynamic planners adapt to changes by selectively repairing precomputed structures. RT-FMT~\cite{Silveira2023RealTimeFM} and RT-RRT$^{*}$~\cite{Naderi2015RTRRTAR} move the root with the agent to handle changing goals in real-time. DRRT~\cite{Ferguson2006ReplanningWR} reverses tree growth (goal-to-robot) and repairs only affected branches. RRT\textsuperscript{X}~\cite{otte2016rrtx} maintains asymptotic optimality in dynamic settings by propagating dynamic-programming-style repair cascades across the graph upon obstacle updates. Lazy Lifelong Planning Tree$^{*}$ (LLPT$^{*}$)~\cite{Huang2025AsymptoticallyOL} extends lazy lifelong planning to continuous spaces using RRG, achieving anytime asymptotic optimality in dynamic environments through efficient graph repair.

FMT\textsuperscript{X} extends FMT$^{*}$ into a dynamic framework, but
differs fundamentally in its use of laziness. While the LazySP paradigm and
continuous extensions such as LLPT$^{*}$ are inherently path-centric,
deferring checks along candidate paths, FMT\textsuperscript{X} roots its
laziness in the wavefront expansion of the Fast Marching method.
It generalizes the discrete unvisited rule of FMT$^{*}$ into a cost-based
consistency condition, replacing a binary visitation decision with a
real-valued test of potential cost suboptimality. The resulting inequality serves as an event-driven
indicator of potential suboptimality and triggers localized Bellman
reevaluations among valid open neighbors only when necessary, thereby
confining structural repairs to corrupted subgraphs.

Crucially, FMT\textsuperscript{X} encompasses two complementary operational
modes. The fixed-graph variant, Dynamic FMT$^{*}$, isolates this efficient
graph-repair mechanism over a fixed batch of pre-sampled vertices.
FMT\textsuperscript{X} extends this mechanism by continuously adding samples
to incrementally refine the space. To support unbounded continuous execution,
FMT\textsuperscript{X} adapts the neighbor-culling and
$\epsilon$-consistency strategies originally introduced in
RRT\textsuperscript{X}~\cite{otte2016rrtx}. The culling mechanism maintains
tractable expected neighborhood sizes as the graph grows, while
$\epsilon$-consistency bounds the depth of cost-propagation cascades.
Combined with a direct parent-to-child propagation step that prevents
cost-propagation failure under FMT$^{*}$'s lazy collision checking, these
mechanisms produce a planner that scales from batch-based dynamic repair to
continuous anytime adaptation.

\section{Problem Statement}
\label{sec:problem_statement}

Let $\mathcal{X}$ denote the robot's state space, equipped with a metric $d$
used for proximity queries. Let
$\mathcal{X}_{\mathrm{obs}}(t)\subset\mathcal{X}$ denote the obstacle region
represented by the planner at time $t$, induced by a finite set of obstacles
$O(t)$, and let $\mathcal{X}_{\mathrm{free}}(t)
=\mathcal{X}\setminus\mathcal{X}_{\mathrm{obs}}(t)$.
The represented region may change with time and with the robot state,
$\Delta\mathcal{X}_{\mathrm{obs}}=f(t,x_{\mathrm{bot}})$, and these changes are
not known a priori. Instead, they are revealed online during execution, and may
arise from physical obstacle motion, from revised predictions of that motion, or
from geometry that becomes observable only as the robot approaches it.

Write $x_{\mathrm{bot}}(t)\in\mathcal{X}$ for the robot state at time $t$. The
robot starts at
$x_{\mathrm{bot}}(0)=x_{\mathrm{start}}\in\mathcal{X}_{\mathrm{free}}(0)$ and
must reach a fixed goal state $x_{\mathrm{goal}}$, assumed to remain in free
space throughout the mission.

A steering trajectory $\pi(x_1,x_2)$ is a continuous curve from $x_1$ to $x_2$
that satisfies the robot's kinematic or kinodynamic constraints. Let $J(\pi)$
denote the cost of a steering-feasible trajectory, assumed additive under
concatenation. Let
\[
d_\pi(x_1,x_2)\;:=\;J\bigl(\pi(x_1,x_2)\bigr),
\]
with $d_\pi(x_1,x_2)=\infty$ when no steering-feasible connection exists. At
time $t$ the connection is collision-free when
$\pi(x_1,x_2)\cap\mathcal{X}_{\mathrm{obs}}(t)=\emptyset$, and it is valid when
it is both steering-feasible and collision-free. Steering feasibility depends on the system and its steering model, whereas
whether a connection is collision-free depends on the current obstacle region.
FMT\textsuperscript{X} exploits that separation by deferring the second test. By additivity, a trajectory composed of successive
steering connections has cost
$J(\pi)=\sum_i d_\pi(x_i,x_{i+1})$. The metric $d$ used for
neighbor queries need not coincide with $d_\pi$.

Given $\mathcal{X}$, $\mathcal{X}_{\mathrm{obs}}(0)$, $x_{\mathrm{goal}}$,
$x_{\mathrm{bot}}(0)=x_{\mathrm{start}}$, and unknown changes
$\Delta\mathcal{X}_{\mathrm{obs}}$, the objective is to find a minimum-cost
valid trajectory from the robot's current state to the goal and, until the
goal is reached, to simultaneously execute the current trajectory and
recompute it whenever new obstacle information changes the planning snapshot:
\begin{equation}
	\label{eq:shortest_replan}
	\pi^*(x_{\mathrm{bot}},x_{\mathrm{goal}})
	=
	\operatorname*{arg\,min}_{\substack{
		\pi(x_{\mathrm{bot}},x_{\mathrm{goal}})\\ \mathrm{valid}}}
	J(\pi).
\end{equation}

At each replanning instant $t_k$, the planner computes
$\pi^*(x_{\mathrm{bot}}(t_k),x_{\mathrm{goal}})$ with respect to the obstacle
information currently available. Planning and execution continue until the
first time $T_{\mathrm{end}}$ such that
$x_{\mathrm{bot}}(T_{\mathrm{end}})=x_{\mathrm{goal}}$.

The geometric, holonomic, Dubins, and thruster models considered in this work
instantiate this common problem through their respective state spaces,
steering functions, feasibility constraints, and edge-cost functionals.
Time-augmented experiments may represent the fixed terminal configuration by
multiple time-indexed goal states, as described in
Section~\ref{subsubsec:replanning_benchmarks}.

\section{Algorithms}
\label{sec:algorithms}

We present two planners based on a unified dynamic graph formulation and a
common lazy wavefront repair mechanism. D-FMT$^{*}$ uses a fixed sample set and
fixed neighborhoods with $\epsilon=0$, whereas FMT\textsuperscript{X} extends the
same machinery to the anytime setting with incremental sampling,
radius-dependent neighborhood maintenance and culling, and
$\epsilon$-consistent queuing. Both planners share the obstacle-update and
wavefront mechanisms below.

\subsection{Notation and dynamic sets}
\label{sec:notation}

Both planners maintain a directed graph $\mathcal{G}=(V,E)$ in the state
space $\mathcal{X}$ and a goal-rooted cost-to-go tree
$\mathcal{T}=(V_{\mathcal{T}},E_{\mathcal{T}})$, with
$V_{\mathcal{T}}\subseteq V$ and $E_{\mathcal{T}}\subseteq E$.
Edges in $E$ are steering-feasible connections.
Let $\pi(x,y)$ be the trajectory returned by steering from $x$ to $y$ when
that connection is feasible. Its edge cost is $d_\pi(x,y)$ as in
Section~\ref{sec:problem_statement}.
Infeasible pairs are omitted from $E$ and from $N^\pm$.
The cost of a path made of successive edges is
$J(\pi)=\sum_i d_\pi(x_i,x_{i+1})$,
so minimizing cost-to-goal in $\mathcal{T}$ approximates
\eqref{eq:shortest_replan}.
For each vertex, $N^{+}(x)$ and $N^{-}(x)$ denote the stored outgoing and
incoming neighborhoods. D-FMT$^{*}$ builds them once from the connections
satisfying $d_\pi\le r_n$ and holds them fixed. FMT\textsuperscript{X} maintains
$N^{\pm}(x)=N_0^{\pm}(x)\cup N_r^{\pm}(x)$, where $N_0^{\pm}$ are the birth
neighborhoods retained permanently and $N_r^{\pm}$ are the radius-dependent
neighborhoods contributed by later insertions and culled as $r_n$ shrinks
(Section~\ref{sec:fmtx}). A radius-dependent link may temporarily exceed the
current $r_n$ until it is culled, while a retained birth link may remain longer
than the current $r_n$. For $A\subseteq V$ we write
$N^{+}(A)\triangleq\bigcup_{x\in A}N^{+}(x)$.
For $x\in V$, $g(x)$ is the last committed cost-to-go label and
$\mathrm{lmc}(x)$ is the tentative look-ahead label maintained by the
wavefront. A neighbor $y\in N^{+}(x)$ with finite $\mathrm{lmc}(y)$ proposes
the candidate value $d_\pi(x,y)+\mathrm{lmc}(y)$. Because collision checking is
lazy, $\mathrm{lmc}(x)$ need not equal the least of these candidates at
intermediate times. For D-FMT$^{*}$, Lemma~\ref{lem:consistency} establishes
the corresponding post-drain inequality, subject to the lazy failed-check
exception stated there.
Labels are initialized to $g=\mathrm{lmc}=\infty$ with $p_{\mathcal{T}}=\emptyset$,
except at the goal, where
$g(x_{\mathrm{goal}})=\mathrm{lmc}(x_{\mathrm{goal}})=0$.
Let $p_{\mathcal{T}}(x)$ and $\mathcal{C}_{\mathcal{T}}(x)$ be the parent and
children of $x$ in $\mathcal{T}$, and let $\hat{x}_{\mathrm{bot}}\in V$ be the
robot's current anchor, a vertex of finite $\mathrm{lmc}$ reachable from the
robot through a steering-feasible, collision-free bridge. Its local selection
policy is an implementation detail and plays no role in the analysis of
Section~\ref{sec:runtime_analysis}.
The priority queue $V_{\mathrm{open}}$ holds nodes ordered by $\mathrm{lmc}$
that may act as parents in the wavefront.
Write $\pi_{\mathcal{T}}(t_k)$ for the trajectory the planner executes at
instant $t_k$, formed by the bridge from the robot to
$\hat{x}_{\mathrm{bot}}$ followed by the tree path to $x_{\mathrm{goal}}$.

When the represented obstacle region $\mathcal{X}_{\mathrm{obs}}$ is updated, tree edges
that intersect the new region are cut.
A node is directly invalidated when its parent edge meets the new obstacles,
\[
\mathcal{V}_{\mathcal{T}}^{\mathrm{inv}}
\triangleq
\bigl\{
x\in V_{\mathcal{T}}\setminus\{x_{\mathrm{goal}}\}
\mid
\pi\bigl(x,p_{\mathcal{T}}(x)\bigr)\cap\mathcal{X}_{\mathrm{obs}}\neq\emptyset
\bigr\},
\]
and the orphan set is that set together with all of its descendants in
$\mathcal{T}$,
\[
\mathcal{V}_{\mathcal{T}}^{\mathrm{orphan}}
\triangleq
\mathcal{V}_{\mathcal{T}}^{\mathrm{inv}}
\cup
\mathrm{Desc}_{\mathcal{T}}\!\bigl(\mathcal{V}_{\mathcal{T}}^{\mathrm{inv}}\bigr).
\]
Each node relies on a single parent edge, so cutting that edge isolates the
node and all of its descendants in $\mathcal{T}$. Detaching the entire orphan
set, rather than only the directly invalidated nodes, is what preserves the
standing invariant used by the analysis of Section~\ref{sec:runtime_analysis}.

\smallskip
\noindent\textbf{(P1) Label soundness.}
Every finite $\mathrm{lmc}$ is backed by a collision-free route to
$x_{\mathrm{goal}}$ in the current snapshot. An obstacle addition sets
$\mathrm{lmc}=g=\infty$ and severs the parent of every node whose current
parent route is invalidated, together with all of its tree descendants, and
modifies no other label.
\smallskip

\noindent
Algorithm~\ref{alg:AddObstacle} detaches every
$x\in\mathcal{V}_{\mathcal{T}}^{\mathrm{orphan}}$ and primes repair by inserting
into $V_{\mathrm{open}}$ every finite-$\mathrm{lmc}$ neighbor of the orphan
boundary. These neighbors are all
$y\in N^{+}(\mathcal{V}_{\mathcal{T}}^{\mathrm{orphan}})$ with
$\mathrm{lmc}(y)<\infty$.

\begin{algorithm}
	\caption{$\mathtt{AddObstacle}(o)$}
	\label{alg:AddObstacle}
	\footnotesize
	\begin{algorithmic}[1]
		\State $\mathcal{V}_{\mathcal{T}}^{\mathrm{inv}}
		\gets \{x \in V_{\mathcal{T}}\setminus\{x_{\mathrm{goal}}\} \mid
		\pi(x,p_{\mathcal{T}}(x)) \cap o \neq \emptyset\}$
		\State $\mathcal{V}_{\mathcal{T}}^{\mathrm{orphan}}
		\gets \mathcal{V}_{\mathcal{T}}^{\mathrm{inv}}
		\cup \mathrm{Desc}_{\mathcal{T}}(\mathcal{V}_{\mathcal{T}}^{\mathrm{inv}})$
		\For{$x \in \mathcal{V}_{\mathcal{T}}^{\mathrm{orphan}}$}
		\If{$x \in V_{\mathrm{open}}$}
		\State $V_{\mathrm{open}}.\mathtt{remove}(x)$
		\EndIf
		\State $\mathrm{lmc}(x) \gets g(x) \gets \infty$
		\State $\mathcal{C}_{\mathcal{T}}(p_{\mathcal{T}}(x))
		\gets \mathcal{C}_{\mathcal{T}}(p_{\mathcal{T}}(x)) \setminus \{x\}$
		\State $p_{\mathcal{T}}(x) \gets \emptyset$
		\EndFor
		\State $V_{\mathrm{open}}
		\gets V_{\mathrm{open}}
		\cup \{y \in N^{+}(\mathcal{V}_{\mathcal{T}}^{\mathrm{orphan}})
		\mid \mathrm{lmc}(y)<\infty\}$
	\end{algorithmic}
\end{algorithm}

When an obstacle $o$ disappears, previously blocked connections may improve
costs.
Let $A_{\mathrm{rem}}(o)$ denote any conservative set containing every vertex
$x$ whose stored outgoing connection may become collision-free once $o$ is
removed, and let $\mathtt{RecoveryNodes}(o)$ denote any set containing
$N^{+}\bigl(A_{\mathrm{rem}}(o)\bigr)$.
A practical broad phase locates the potentially affected vertices near $o$ and
then examines their stored neighborhoods. In the geometric setting, if $o$ is
contained in a ball of
radius $R_o$ and $\rho$ upper-bounds the length of any retained edge, then
querying the vertices within $R_o+\rho$ of that ball's center locates every
affected endpoint, since a newly freed edge meets $o$ and has length at most
$\rho$. In D-FMT$^{*}$ one may take $\rho=r_n$, and in
FMT\textsuperscript{X} one may take $\rho=\delta$ by the radius cap of
Section~\ref{sec:fmtx}.
Together with the orphan-boundary seeding above, this gives the second property
used by Section~\ref{sec:runtime_analysis}.

\begin{algorithm}
	\caption{$\mathtt{RemoveObstacle}(o)$}
	\label{alg:RemoveObstacle}
	\footnotesize
	\begin{algorithmic}[1]
		\State $\mathcal{V}^{\mathrm{rec}} \gets \mathtt{RecoveryNodes}(o)$
		\State $V_{\mathrm{open}}
		\gets V_{\mathrm{open}}
		\cup \{x\in\mathcal{V}^{\mathrm{rec}} \mid \mathrm{lmc}(x)<\infty\}$
	\end{algorithmic}
\end{algorithm}

\smallskip
\noindent\textbf{(P2) Seed completeness.}
Let $S$ be the set of vertices an update inserts into $V_{\mathrm{open}}$. For
an obstacle addition let
$A_{\mathrm{add}}=\mathcal{V}_{\mathcal{T}}^{\mathrm{orphan}}$, and for an
obstacle removal let $A_{\mathrm{rem}}$ contain every vertex $x$ for which some
stored outgoing connection $(x,u)$ may become collision-free as a result of the
removal. With $A=A_{\mathrm{add}}\cup A_{\mathrm{rem}}$, every $x\in A$ and
every $y\in N^{+}(x)$ with $\mathrm{lmc}(y)<\infty$ satisfies $y\in S$. The
geometric recovery procedure described above is one conservative
implementation of this requirement.
\smallskip

\noindent
Orphans are reconnected passively when a seeded parent is extracted and
relaxes its incoming neighbors. Since \textup{(P2)} applies only to neighbors
with $\mathrm{lmc}(y)<\infty$, only the boundary of the orphan set is seeded,
and the wavefront then advances inward.

Algorithm~\ref{alg:queuenode} inserts or reprioritizes nodes.
If $x\in V_{\mathrm{open}}$, its key becomes $\mathrm{lmc}(x)$.
Otherwise $x$ is inserted only when $g(x)-\mathrm{lmc}(x)>\epsilon$.
D-FMT$^{*}$ always uses $\epsilon=0$. FMT\textsuperscript{X} uses $\epsilon>0$
(Section~\ref{sec:fmtx}).

\begin{algorithm}
	\caption{$\mathtt{QueueNode}(x,\epsilon)$}
	\label{alg:queuenode}
	\footnotesize
	\begin{algorithmic}[1]
		\If{$x \in V_{\mathrm{open}}$}
		\State $V_{\mathrm{open}}.\mathtt{update}(x,\mathrm{lmc}(x))$
		\ElsIf{$g(x)-\mathrm{lmc}(x)>\epsilon$}
		\State $V_{\mathrm{open}}.\mathtt{insert}(x,\mathrm{lmc}(x))$
		\EndIf
	\end{algorithmic}
\end{algorithm}

\subsection{Wavefront}
\label{sec:expand}

Algorithm~\ref{alg:Expand} is the cost-ordered expansion used by both
planners.
Classical FMT$^{*}$ expands through an unvisited set
$V_{\mathrm{unvisited}}$~\cite{janson2015fast}.
Here a neighbor of an extracted node $z$ is considered only under the
cost test $\mathrm{lmc}(x)>g(z)+d_\pi(x,z)$.
That criterion is what allows the same loop to
repair $\mathcal{T}$ after a dynamic change. The finite-cost neighbors of orphaned
subtrees and the recovery nodes are seeded into $V_{\mathrm{open}}$, and any node that is still
suboptimal through an extracted neighbor $z$ can be considered again.
After $g(z)\gets\mathrm{lmc}(z)$, the body has two roles (detailed in the
listing):

\begin{itemize}
	\item \textbf{Tree propagation.}
	Push improvements to every child $x\in\mathcal{C}_{\mathcal{T}}(z)$ along
	the known-valid tree edge (no collision checks), then
	$\mathtt{QueueNode}(x,\epsilon)$.
	This direct cost push from each parent to its children prevents
	cost-propagation failure when a candidate parent connection fails its lazy
	collision check or when, in FMT\textsuperscript{X}, culling removes a
	non-birth geometric neighbor that the wavefront would otherwise have used, so
	subtrees do not keep stale $\mathrm{lmc}$ values.
	
	\item \textbf{Implicit rewiring.}
	Among $x\in N^-(z)$ still suboptimal through $z$, choose a best parent in
	$N^+(x)\cap V_{\mathrm{open}}$ and accept it if it is already
	$p_{\mathcal{T}}(x)$ or $\mathtt{CollisionFree}$.
	Restricting candidates to $V_{\mathrm{open}}$ keeps parent selection on the
	active wavefront.
\end{itemize}

The loop runs while $V_{\mathrm{open}}$ is nonempty and the best queued
$\mathrm{lmc}$ can still improve $\hat{x}_{\mathrm{bot}}$, bounding work per tick.
Section~\ref{sec:runtime_analysis} uses only the following mechanics of this
loop.

\smallskip
\noindent\textbf{(P3) Wavefront discipline.}
Within a snapshot every write to a finite $\mathrm{lmc}$ is a strict
decrease. Every such decrease at an existing vertex is passed to
$\mathtt{QueueNode}(\cdot,\epsilon)$, which decrease-keys the vertex when it
is already open and inserts it when $g-\mathrm{lmc}>\epsilon$, and every newly
admitted vertex of finite $\mathrm{lmc}$ is inserted directly into
$V_{\mathrm{open}}$ by $\mathtt{Extend}$. Under
$\epsilon=0$, as in D-FMT$^{*}$, every decrease therefore enters
$V_{\mathrm{open}}$. On extraction, $z$ sets $g(z)\gets\mathrm{lmc}(z)$ and
performs both relaxations above, the second being the FMT$^{*}$ connection
step~\cite{janson2015fast} restricted to candidates in $V_{\mathrm{open}}$.
\smallskip

Algorithm~\ref{alg:Expand} is stated in the general case used by FMT\textsuperscript{X},
including calls to $\mathtt{CullNeighbors}$.
D-FMT$^{*}$ uses fixed neighborhoods $N^\pm$ built at initialization and
sets $\epsilon=0$. With fixed $r_n$, every stored edge already satisfies
$d_\pi\le r_n$, so $\mathtt{CullNeighbors}$ is a no-op.
FMT\textsuperscript{X} maintains $N^\pm\equiv N_0^\pm\cup N_r^\pm$, runs culling as $r_n$
shrinks, and uses $\epsilon>0$
(Section~\ref{sec:fmtx}).

\begin{algorithm}
	\caption{$\mathtt{Expand}()$}
	\label{alg:Expand}
	\footnotesize
	\begin{algorithmic}[1]
		\While{%
			\parbox[t]{0.78\linewidth}{%
				$V_{\mathrm{open}}\neq\emptyset$ \textbf{and}
				$\bigl($
				$\mathrm{lmc}(V_{\mathrm{open}}.\mathtt{top}())<\mathrm{lmc}(\hat{x}_{\mathrm{bot}})$\\
				\hspace*{1.2em}\textbf{or}
				$\hat{x}_{\mathrm{bot}}\in V_{\mathrm{open}}$
				\textbf{or}
				$\mathrm{lmc}(\hat{x}_{\mathrm{bot}})=\infty$
				$\bigr)$
				\hspace*{0.25em} \textbf{do}
		}}
		\State $z \gets V_{\mathrm{open}}.\mathtt{top}()$
		\State $g(z) \gets \mathrm{lmc}(z)$
		\For{$x \in \mathcal{C}_{\mathcal{T}}(z)$}
		\If{$\mathrm{lmc}(x) > g(z)+d_\pi(x,z)$}
		\State $\mathrm{lmc}(x) \gets g(z)+d_\pi(x,z)$
		\State $\mathtt{QueueNode}(x,\epsilon)$
		\EndIf
		\EndFor
		\State $\mathtt{CullNeighbors}(z,r_n)$
		\For{$x \in \{x'\in N^{-}(z)\mid \mathrm{lmc}(x')>g(z)+d_\pi(x',z)\}$}
		\State $\mathtt{CullNeighbors}(x,r_n)$
		\State $Y_{\mathrm{better}} \gets \{y\in N^{+}(x)\cap V_{\mathrm{open}}
		\mid \mathrm{lmc}(y)+d_\pi(x,y)<\mathrm{lmc}(x)\}$
		\If{$Y_{\mathrm{better}}\neq\emptyset$}
		\State $y_{\min} \gets \arg\min_{y\in Y_{\mathrm{better}}}
		\bigl(\mathrm{lmc}(y)+d_\pi(x,y)\bigr)$
		\If{$y_{\min}=p_{\mathcal{T}}(x)$ \textbf{or}
			$\mathtt{CollisionFree}(x,y_{\min})$}
		\State $\mathrm{lmc}(x)
		\gets \mathrm{lmc}(y_{\min})+d_\pi(x,y_{\min})$
		\If{$y_{\min}\neq p_{\mathcal{T}}(x)$}
		\If{$p_{\mathcal{T}}(x)\neq\emptyset$}
		\State $\mathcal{C}_{\mathcal{T}}(p_{\mathcal{T}}(x)) \gets
		\mathcal{C}_{\mathcal{T}}(p_{\mathcal{T}}(x))\setminus\{x\}$
		\EndIf
		\State $p_{\mathcal{T}}(x) \gets y_{\min}$
		\State $\mathcal{C}_{\mathcal{T}}(y_{\min}) \gets
		\mathcal{C}_{\mathcal{T}}(y_{\min})\cup\{x\}$
		\EndIf
		\State $\mathtt{QueueNode}(x,\epsilon)$
		\EndIf
		\EndIf
		\EndFor
		\State $V_{\mathrm{open}} \gets V_{\mathrm{open}}\setminus\{z\}$

		\EndWhile
	\end{algorithmic}
\end{algorithm}
\subsection{D-FMT$^{*}$}
\label{sec:dfmt}

D-FMT$^{*}$ (Algorithm~\ref{alg:DFMTStar}) is the fixed-batch form of the same
cost-consistency wavefront. It keeps $V$ and $N^\pm$ fixed, sets
$\epsilon=0$, and uses implicit rewiring with parent-to-child propagation to
repair $\mathcal{T}$ when obstacles move, without densifying the
graph or restarting FMT$^{*}$~\cite{janson2015fast}.
It samples $n$ states, builds $N^\pm$ at radius $r_n$, and initializes
$V_{\mathrm{open}}\gets\{x_{\mathrm{goal}}\}$.
At each $t_k$, if the current prediction differs from the represented obstacle set $O$,
$\mathtt{UpdateObstacles}$ applies removals then additions, seeding recovery
nodes and orphaning $\mathcal{V}_{\mathcal{T}}^{\mathrm{orphan}}$ as above.
$\mathtt{Expand}$ is then called to repair $\mathcal{T}$ before the robot tracks $\pi_{\mathcal{T}}(t_k)$.

\begin{algorithm}
	\caption{$\mathtt{UpdateObstacles}(O^{\mathrm{new}})$}
	\label{alg:UpdateObstacles}
	\footnotesize
	\begin{algorithmic}[1]
		\State $O^{\mathrm{old}} \gets O$
		\For{$o \in O^{\mathrm{old}} \setminus O^{\mathrm{new}}$}
		\State $\mathtt{RemoveObstacle}(o)$
		\EndFor
		\For{$o \in O^{\mathrm{new}} \setminus O^{\mathrm{old}}$}
		\State $\mathtt{AddObstacle}(o)$
		\EndFor
		\State $O \gets O^{\mathrm{new}}$
	\end{algorithmic}
\end{algorithm}

The key idea is localized repair. Every tree segment that remains valid under
the updated obstacle configuration is preserved, and $\mathtt{Expand}$ is
started only from orphan boundaries and recovery nodes.
Finite $\mathrm{lmc}$ values on surviving tree segments remain feasible upper
bounds on the true cost-to-goal.
They may already be tight, or may be conservative overestimates that tree
propagation and subsequent parent re-evaluation later tighten.

D-FMT$^{*}$ keeps every still-valid parent link across obstacle updates and
only reopens orphan boundaries and recovery nodes.
$\mathtt{Expand}$ can then tighten $\mathrm{lmc}$ on nodes that are already
in $\mathcal{T}$, rather than discarding the tree and rebuilding it on all
of $V$.
Under the assumptions of Section~\ref{sec:runtime_analysis}, the repaired cost
satisfies the same asymptotic optimality guarantee as FMT$^{*}$. Equality of the
finite-$n$ trees is neither claimed nor required.
With $V$ fixed, that batch quality is the target. Anytime densification is
left to FMT\textsuperscript{X}.

\begin{algorithm}
	\caption{Dynamic FMT$^{*}$ (D-FMT$^{*}$)}
	\label{alg:DFMTStar}
	\footnotesize
	\begin{algorithmic}[1]
		\State $V \gets \{x_{\mathrm{bot}}(0), x_{\mathrm{goal}}\} \cup \mathtt{Sample}(n)$
		\State $V_{\mathrm{open}} \gets \{x_{\mathrm{goal}}\}$
		\State $O \gets \mathtt{GetObstaclePrediction}(0)$
		\State $r_n \gets \mathtt{ComputeRadius}(|V|)$
		\State $\epsilon \gets 0$
		\For{$x \in V$}
		\State $N^+(x) \gets \{y \in V \setminus \{x\} \mid d_\pi(x,y) \le r_n\}$
		\State $N^-(x) \gets \{y \in V \setminus \{x\} \mid d_\pi(y,x) \le r_n\}$
		\EndFor
		\While{$x_{\mathrm{bot}}(t_k)\neq x_{\mathrm{goal}}$}
		\State $O^{\mathrm{new}} \gets \mathtt{GetObstaclePrediction}(t_k)$
		\If{$O^{\mathrm{new}} \neq O$}
		\State $\mathtt{UpdateObstacles}(O^{\mathrm{new}})$
		\EndIf
		\State $\mathtt{Expand}()$
		\State $x_{\mathrm{bot}}(t_{k+1}) \gets \mathtt{Execute}(\pi_{\mathcal{T}}(t_k), \Delta t)$
		\State $t_{k+1} \gets t_k + \Delta t$
		\EndWhile
	\end{algorithmic}
\end{algorithm}

\subsection{FMT\textsuperscript{X}}
\label{sec:fmtx}

FMT\textsuperscript{X} (Algorithm~\ref{alg:fmtx}) keeps the same
$\mathtt{UpdateObstacles}$ and the same $\mathtt{Expand}$, and grows
$V$ for anytime improvement using $\epsilon$-consistency and radius culling
in the spirit of RRT\textsuperscript{X}~\cite{otte2016rrtx}.
From $\{x_{\mathrm{bot}}(0),x_{\mathrm{goal}}\}$, batches of $m$ samples enter
via $\mathtt{Extend}$ (Algorithm~\ref{alg:Extend}).
Each sample is steered into the graph, attached to a best collision-free
finite-$\mathrm{lmc}$ parent, and on success inserted into $V$
and $V_{\mathrm{open}}$.

\begin{algorithm}
	\caption{\texorpdfstring{$\mathtt{Extend}(m)$}{Extend(m)}}
	\label{alg:Extend}
	\footnotesize
	\begin{algorithmic}[1]
		\For{$i=1,\ldots,m$}
		\State $x_{\mathrm{rand}}
		\gets \mathtt{Sample}(\mathcal{X}_{\mathrm{free}})$
		\State $x_{\mathrm{near}}
		\gets \mathtt{Nearest}(V,x_{\mathrm{rand}})$
		\State $x_{\mathrm{new}}
		\gets \mathtt{Saturate}(x_{\mathrm{rand}},x_{\mathrm{near}},\delta)$
		\State $V_{\mathrm{near}}
		\gets\{y\in V\mid d(x_{\mathrm{new}},y)\le r_n\}$
		\State $Y \gets \{y\in V_{\mathrm{near}} \mid
		d_\pi(x_{\mathrm{new}},y)\le r_n \land \mathrm{lmc}(y)<\infty\}$
		\State $\mathrm{lmc}(x_{\mathrm{new}})\gets\infty$
		\While{$Y\neq\emptyset$}
		\State $y^* \gets \arg\min_{y\in Y}
		\bigl(\mathrm{lmc}(y)+d_\pi(x_{\mathrm{new}},y)\bigr)$
		\If{$\mathtt{CollisionFree}(x_{\mathrm{new}},y^*)$}
		\State $\mathrm{lmc}(x_{\mathrm{new}})
		\gets \mathrm{lmc}(y^*)+d_\pi(x_{\mathrm{new}},y^*)$
		\State $p_{\mathcal{T}}(x_{\mathrm{new}}) \gets y^*$
		\State $\mathcal{C}_{\mathcal{T}}(y^*)\gets\mathcal{C}_{\mathcal{T}}(y^*)\cup\{x_{\mathrm{new}}\}$
		\State \textbf{break}
		\EndIf
		\State $Y\gets Y\setminus\{y^*\}$
		\EndWhile
		\If{$\mathrm{lmc}(x_{\mathrm{new}})<\infty$}
		\ForAll{$y\in V_{\mathrm{near}}$}
		\If{$d_\pi(x_{\mathrm{new}},y)\le r_n$}
		\State $N_0^+(x_{\mathrm{new}})
		\gets N_0^+(x_{\mathrm{new}})\cup\{y\}$
		\State $N_r^-(y)
		\gets N_r^-(y)\cup\{x_{\mathrm{new}}\}$
		\EndIf
		\If{$d_\pi(y,x_{\mathrm{new}})\le r_n$}
		\State $N_0^-(x_{\mathrm{new}})
		\gets N_0^-(x_{\mathrm{new}})\cup\{y\}$
		\State $N_r^+(y)
		\gets N_r^+(y)\cup\{x_{\mathrm{new}}\}$
		\EndIf
		\EndFor
		\State $V\gets V\cup\{x_{\mathrm{new}}\}$
		\State $V_{\mathrm{open}}
		\gets V_{\mathrm{open}}\cup\{x_{\mathrm{new}}\}$
		\EndIf
		\EndFor
		\State $r_n\gets\mathtt{ComputeRadius}(|V|)$
	\end{algorithmic}
\end{algorithm}

These two steps mirror RRT$^{*}$~\cite{karaman2011sampling}. Attaching a new
sample to its best collision-free parent within $r_n$ is RRT$^{*}$'s best-parent
selection, and the sweep of $N^{-}$ performed when that sample is later
extracted plays the role of RRT$^{*}$'s rewiring of the same ball. What differs
is when the second step runs. RRT$^{*}$ rewires eagerly at insertion, whereas
FMT\textsuperscript{X} defers it to the cost-ordered wavefront. That deferral is
what preserves lazy collision checking, and it is also what admits the
finite-sample suboptimal-connection event of FMT$^{*}$. The
asymptotic-optimality guarantee established in
Section~\ref{subsec:dfmt_ao} is convergence in probability.

To control degree as $|V|$ grows, FMT\textsuperscript{X} distinguishes birth
neighborhoods $N_0^\pm$, recorded when a sample is inserted, from the
radius-dependent neighborhoods maintained online.
At insertion, each new node is linked into $N_0^\pm$ among nodes within the
current $r_n$.
After each batch, $r_n$ is updated from $|V|$ and capped by the
steering-extension bound,
\[
r_n=\min\Bigl\{\,\delta,\;C\gamma\bigl(\log n/n\bigr)^{1/d}\Bigr\},
\]
where $d\ge2$ denotes the fixed dimension used in the neighborhood-radius
scaling,
so every edge has birth length at most $\delta$ and $\rho=\delta$ upper-bounds
the length of any retained edge.
$\mathtt{CullNeighbors}$ (Algorithm~\ref{alg:cullneighbors}) may delete a
stored edge only if it is longer than the current $r_n$, is not the tree
parent, and is not a birth edge. Members of $N_0^\pm$ are never
culled. As in RRT\textsuperscript{X}, culling may make the locally stored
neighborhoods asymmetric, since each vertex retains its own birth neighbors
while reciprocal radius-dependent records at the opposite endpoint may later be
removed.

\begin{algorithm}
	\caption{$\mathtt{CullNeighbors}(v,r_n)$}
	\label{alg:cullneighbors}
	\footnotesize
	\begin{algorithmic}[1]
		\ForAll{$u\in N_r^+(v)$}
		\If{$d_\pi(v,u)>r_n$ \textbf{and} $p_{\mathcal{T}}(v)\neq u$}
		\State $N_r^+(v)\gets N_r^+(v)\setminus\{u\}$
		\State $N_r^-(u)\gets N_r^-(u)\setminus\{v\}$
		\EndIf
		\EndFor
	\end{algorithmic}
\end{algorithm}
Non-birth edges that remain within the shrinking ball stay available as
radius-dependent neighbors.
Birth edges are retained for connectivity and optimality alongside those
surviving non-birth edges.

Every tick runs $\mathtt{Extend}(m)$ then $\mathtt{Expand}$, and applies
obstacle repair when $O$ changes.
$\mathtt{QueueNode}(\cdot,\epsilon)$ ignores inconsistencies at most
$\epsilon$.

Tree propagation remains mandatory under culling. A long non-birth
geometric edge may leave the radius-dependent neighborhoods even though
children in $\mathcal{C}_{\mathcal{T}}$ must stay cost-consistent, and a lazy
check may reject a rewire.
Birth neighbors $N_0^\pm$ remain available to their owner's later parent
search.
The parent-to-child push keeps $\mathcal{T}$ cost-consistent in these cases
and prevents cost-propagation failure when lazy rewiring does not accept a
new parent.

The stop rule matches D-FMT$^{*}$. Continued sampling provides anytime
refinement, and Section~\ref{sec:runtime_analysis} proves asymptotic optimality
for the geometric, $\epsilon=0$, full-drain variant.
As in RRT\textsuperscript{X}, $\epsilon>0$ is an anytime consistency tolerance on
when inconsistent nodes re-enter $V_{\mathrm{open}}$.

\begin{algorithm}
	\caption{FMT\textsuperscript{X}}
	\label{alg:fmtx}
	\footnotesize
	\begin{algorithmic}[1]
		\State $V \gets \{x_{\mathrm{bot}}(0),x_{\mathrm{goal}}\}$
		\State $V_{\mathrm{open}}\gets\{x_{\mathrm{goal}}\}$
		\State $O\gets\mathtt{GetObstaclePrediction}(0)$
		\State $r_n\gets\mathtt{ComputeRadius}(|V|)$
		\State $\epsilon \gets \epsilon_0$
		\While{$x_{\mathrm{bot}}(t_k)\neq x_{\mathrm{goal}}$}
		\State $O^{\mathrm{new}}\gets\mathtt{GetObstaclePrediction}(t_k)$
		\If{$O^{\mathrm{new}}\neq O$}
		\State $\mathtt{UpdateObstacles}(O^{\mathrm{new}})$
		\EndIf
		\State $\mathtt{Extend}(m)$
		\State $\mathtt{Expand}()$
		\State $x_{\mathrm{bot}}(t_{k+1})
		\gets \mathtt{Execute}(\pi_{\mathcal{T}}(t_k),\Delta t)$
		\State $t_{k+1}\gets t_k+\Delta t$
		\EndWhile
	\end{algorithmic}
\end{algorithm}

\section{Theoretical Analysis}
\label{sec:runtime_analysis}

This section establishes asymptotic optimality for D-FMT$^{*}$ and
FMT\textsuperscript{X} separately.

D-FMT$^{*}$ operates on a fixed sample set. For each static planning
interval following a map change, we adapt the covering-chain argument of
FMT$^{*}$~\cite{janson2015fast}. In particular,
Lemma~\ref{lem:cost-equivalence} identifies the static obstacle-free
special case in which the local cost-improvement test coincides exactly
with the unvisited-neighbor rule of classical FMT$^{*}$.
The D-FMT$^{*}$ repair argument is expressed through the three properties
\textup{(P1)}--\textup{(P3)} established in Section~\ref{sec:algorithms}, so it
is independent of any particular implementation.
For D-FMT$^{*}$, Lemma~\ref{lem:consistency} then shows that a completed drain
leaves the graph locally consistent, which reduces the covering-chain bound to a telescoping
sum in Lemma~\ref{lem:path-bound} and yields Theorem~\ref{thm:dfmt-ao}, and
Corollary~\ref{cor:mission-ao} extends it uniformly over a mission.

FMT\textsuperscript{X} grows the sample set online.
Corollary~\ref{cor:fmtx-ao} relies on the standard asymptotically optimal graph
construction of RRT$^{*}$/RRT\textsuperscript{X} for the existence of an
asymptotically optimal chain, and shows that the lazy cost-ordered wavefront does
not destroy its cost bound.

\subsection{Equivalence of neighbor-selection rules}
\label{subsec:equivalence-neighbor-selection}

Lemma~\ref{lem:cost-equivalence} concerns forward FMT$^{*}$.
The cost-to-come $c(\cdot)$ is measured from $x_{\mathrm{init}}$, nodes
are extracted from $V_{\mathrm{open}}$ in nondecreasing order of
$c(\cdot)$, and neighborhoods are those of the $r_n$-disk graph.
The setting is static and obstacle-free,
$\mathcal{X}_{\mathrm{free}}=\mathcal{X}$.
Consequently, every disk-graph edge is collision-free and the Bellman
updates performed by FMT$^{*}$ coincide with shortest-path updates on the
$r_n$-disk graph.

\begin{lemma}[Equivalence of neighbor-selection rules]
	\label{lem:cost-equivalence}
	Run forward FMT$^{*}$ on a fixed sample set $V$ in an obstacle-free
	environment with connection radius $r_n>0$. Let
	$V_{\mathrm{open}}$, $V_{\mathrm{closed}}$, and
	$V_{\mathrm{unvisited}}
	=V\setminus(V_{\mathrm{open}}\cup V_{\mathrm{closed}})$ be the standard
	FMT$^{*}$ partition, with $c(x)=+\infty$ for
	$x\in V_{\mathrm{unvisited}}$. For an extracted node
	$z\in V_{\mathrm{open}}$, let
	\[
	N_z=\{x\in V\setminus\{z\}\mid \|x-z\|<r_n\},
	\qquad
	X_{\mathrm{unvisited}}(z)=N_z\cap V_{\mathrm{unvisited}},
	\]
	and
	\[
	X_{\mathrm{cost}}(z)
	=
	\{x\in N_z\mid c(x)>c(z)+\mathrm{Cost}(z,x)\}.
	\]
	Then $X_{\mathrm{cost}}(z)=X_{\mathrm{unvisited}}(z)$.
\end{lemma}

\begin{proof}
	We prove both inclusions.
	
	\begin{enumerate}[label=\textup{(\roman*)}, leftmargin=*]
		
		\item Let $x\in X_{\mathrm{unvisited}}(z)$. Then $x\in N_z$ and
		$c(x)=+\infty$. Since the extracted node $z$ has finite cost,
		$c(x)>c(z)+\mathrm{Cost}(z,x)$. Hence
		$x\in X_{\mathrm{cost}}(z)$, which proves
		$X_{\mathrm{unvisited}}(z)\subseteq X_{\mathrm{cost}}(z)$.
		
		\item Let $x\in X_{\mathrm{cost}}(z)$. Then $x\in N_z$ and
		\begin{equation}
			\label{eq:cost-strict-local}
			c(x)>c(z)+\mathrm{Cost}(z,x).
		\end{equation}
		Assume, for contradiction, that
		$x\in V_{\mathrm{open}}\cup V_{\mathrm{closed}}$. Then $x$ has already
		been discovered. The induction establishing Invariant~2 in the proof of
		Theorem~3.2 of Janson et~al.~\cite{janson2015fast} implies that every
		discovered node in obstacle-free FMT$^{*}$ has its shortest-path cost in
		the $r_n$-disk graph. Since classical FMT$^{*}$ never changes a label
		after discovery, $c(v)=c^{*}(v)$ for every
		$v\in V_{\mathrm{open}}\cup V_{\mathrm{closed}}$, where $c^{*}(v)$ is
		the shortest-path cost from $x_{\mathrm{init}}$ to $v$.
		
		In particular, $c(x)=c^{*}(x)$ and $c(z)=c^{*}(z)$. Because
		$x\in N_z$, $(z,x)$ is a disk-graph edge, so a shortest path to $z$
		followed by $(z,x)$ is a candidate path to $x$. Therefore,
		$c(x)=c^{*}(x)\le c^{*}(z)+\mathrm{Cost}(z,x)
		=c(z)+\mathrm{Cost}(z,x)$, contradicting
		\eqref{eq:cost-strict-local}. Thus,
		$x\notin V_{\mathrm{open}}\cup V_{\mathrm{closed}}$, so
		$x\in V_{\mathrm{unvisited}}$. Together with $x\in N_z$, this gives
		$x\in X_{\mathrm{unvisited}}(z)$, proving
		$X_{\mathrm{cost}}(z)\subseteq X_{\mathrm{unvisited}}(z)$.
		
	\end{enumerate}
	
	Therefore, $X_{\mathrm{cost}}(z)=X_{\mathrm{unvisited}}(z)$.
\end{proof}

\begin{remark}
	\label{rem:cost-static}
	Lemma~\ref{lem:cost-equivalence} shows that the local cost-improvement
	test introduces no revisits in static obstacle-free FMT$^{*}$. A node
	that has already received its exact $r_n$-disk-graph shortest-path label
	cannot satisfy the strict cost-improvement inequality from any adjacent
	extracted node.
	
	The test also supplies a local, stateless criterion that does not
	require an explicit set $V_{\mathrm{unvisited}}$. This criterion remains
	well-defined after obstacle updates and incremental sampling, where the
	one-pass unvisited-set interpretation of classical FMT$^{*}$, and hence
	rebuilding $V_{\mathrm{unvisited}}$, is no longer natural. It underpins
	the wavefront repair of Section~\ref{sec:expand} and the anytime
	densification of Section~\ref{sec:fmtx}.
\end{remark}

\subsection{Goal-rooted interpretation and obstacle effects}
\label{subsec:goal-rooted-interpretation}

With the goal-rooted notation of Section~\ref{sec:notation}, the forward
test in Lemma~\ref{lem:cost-equivalence} is precisely the predicate used
by $\mathtt{Expand}$:
\begin{equation}
	\label{eq:lmc_consistency}
	\mathrm{lmc}(x)>g(z)+d_\pi(x,z),
	\qquad
	x\in N^{-}(z).
\end{equation}
On a static obstacle-free batch, if every finite label equals the exact
disk-graph cost-to-goal, then~\eqref{eq:lmc_consistency} holds exactly for
nodes with $\mathrm{lmc}(x)=+\infty$. Thus, it is the goal-rooted analog
of the unvisited-neighbor rule and cannot select an already optimally
labeled node.

With obstacles, the exact-label property need not hold. FMT$^{*}$ can
commit to a suboptimal parent when the locally best open connection is
blocked while a better parent has already left $V_{\mathrm{open}}$.
Janson et~al.~\cite{janson2015fast} identify four simultaneous geometric
conditions for this lazy suboptimal-connection event. The relevant samples
lie within the connection radius of an obstacle, and the fraction of such
samples vanishes as $n\to\infty$.

An obstacle-induced revisit of an already connected node $x$ requires
more than such a suboptimal connection. If $x$ had its exact
collision-free disk-graph label, the argument of
Lemma~\ref{lem:cost-equivalence} would preclude every strict cost improvement
through an adjacent extracted node. Hence, a revisit first requires that
the lazy obstacle handling leave $x$ with a nonoptimal finite label. It
then additionally requires a later adjacent extracted node $z'$ satisfying
$g(z')+d_\pi(x,z')<\mathrm{lmc}(x)$. Thus, obstacle-induced revisits can
arise only through the same obstacle-local loss of exact-label optimality
that produces lazy suboptimal connections. For this obstacle-induced
mechanism, the relevant nodes lie within the $r_n$-neighborhood of an
obstacle boundary. Consequently, the expected fraction of nodes eligible
for obstacle-induced revisits vanishes as $n\to\infty$.

After orphaning or recovery in D-FMT$^{*}$ or FMT\textsuperscript{X}, a
firing of~\eqref{eq:lmc_consistency} instead triggers intentional
wavefront repair. These repair firings are distinct from
obstacle-induced revisits.

\subsection{Asymptotic optimality of D-FMT$^{*}$ and FMT\textsuperscript{X}}
\label{subsec:dfmt_ao}

We prove asymptotic optimality for geometric D-FMT$^{*}$ on static repair
intervals, and FMT\textsuperscript{X} follows as a corollary. Costs are goal-rooted, so the tree
is rooted at $x_{\mathrm{goal}}$ and the returned cost for a query
$x_{\mathrm{init}}$ is the repaired lookahead
$c_n=\mathrm{lmc}(x_{\mathrm{init}})$.

A dynamic planner built on an eager planner can inherit its optimality
by construction. RRT\textsuperscript{X} transfers asymptotic optimality from
RRT$^{*}$ through its retained original-neighbor structure, which keeps an
RRT$^{*}$ solution realizable as a subgraph of the search
graph~\cite{otte2016rrtx,karaman2011sampling}. Laziness removes that
shortcut. FMT\textsuperscript{X} cannot use the argument directly, because its lazy
parent selection may reject the currently preferred connection and so produce
the finite-sample suboptimal-connection event inherited from FMT$^{*}$. Repair may preserve validated edges that a from-scratch run would skip under the
suboptimal-connection event of~\cite[Sec.~3.2]{janson2015fast}, while a
from-scratch run may make connections absent from the repaired tree. Their
finite-$n$ trees therefore need not agree. Instead, the lazy discrepancy becomes
asymptotically irrelevant as the suboptimal-connection event vanishes.

For D-FMT$^{*}$, the probabilistic content remains that of FMT$^{*}$
\cite[Thm.~4.1, Lem.~4.2--4.4]{janson2015fast}, which we use directly. The
covering-ball construction is not reproved here. What must be added is a
characterization of the labeling a repair leaves behind, since unlike
FMT$^{*}$ we do not start from a trivial initial state. That characterization
is local consistency. Lemma~\ref{lem:consistency} shows a completed drain
leaves every stored edge locally consistent, the seeding rule being exactly
what closes the cases the obstacle update would otherwise break, with a single
exception inherited from FMT$^{*}$'s lazy connection step.
The exception cannot fire on the covering chain. Within an interval this is
FMT$^{*}$'s own clearance argument \textup{(C2)}, and across intervals it is
Lemma~\ref{lem:no-defer}, since an obstacle responsible for an earlier one
cannot have departed without its removal reseeding the whole neighborhood. Given
that invariant, the covering-chain bound is a
telescoping sum (Lemma~\ref{lem:path-bound}) and the theorem follows.

\paragraph{Standing hypotheses.}
\begin{enumerate}[label=\textup{(H\arabic*)},leftmargin=*]
	\item \textbf{FMT$^{*}$ geometry.}
	As in Theorem~4.1 of~\cite{janson2015fast}, with $n$ samples, radius $r_n$,
	and weak $\delta$-clearance, so that $d_\pi$ is the Euclidean distance. In
	D-FMT$^{*}$ the neighborhoods of Section~\ref{sec:notation} are Euclidean
	$r_n$-balls. FMT\textsuperscript{X} instead uses the graph with retained
	birth neighborhoods described in Section~\ref{sec:algorithms}.
	\item \textbf{Full-drain intervals.}
	The stop rule in Algorithm~\ref{alg:Expand} bounds per-tick work during
	execution and is disabled throughout the analysis, since asymptotic
	optimality is a property of the labeling on $V$ rather than of the executed
	trajectory. We call the resulting exhaustive expansion a full drain,
	and every static interval is assumed to end only after
	$V_{\mathrm{open}}=\emptyset$.
	The first begins from the initial labeling, in which
	$\mathrm{lmc}(x_{\mathrm{goal}})=g(x_{\mathrm{goal}})=0$, while
	$\mathrm{lmc}(v)=g(v)=\infty$ and $p_{\mathcal{T}}(v)=\emptyset$ for
	$v\neq x_{\mathrm{goal}}$, and $V_{\mathrm{open}}=\{x_{\mathrm{goal}}\}$.
	Each later interval begins from the terminal state of the preceding repair.
	\item \textbf{Root.}
	$g(x_{\mathrm{goal}})=\mathrm{lmc}(x_{\mathrm{goal}})=0$ at all times, and the root
	is never orphaned.
\end{enumerate}

\paragraph{Covering chain for D-FMT$^{*}$.}
We use the covering-ball construction of~\cite[Thm.~4.1]{janson2015fast} on the
current free snapshot. Let $G_n$ be its good event, that every covering ball
contains a sample and all but a vanishing fraction of the inner balls do, and
let $L_n$ be the accompanying length bound, with $P(G_n^c)\to0$
by~\cite[Lem.~4.3--4.4]{janson2015fast}. Read from $x_{\mathrm{goal}}$ to
$x_{\mathrm{init}}$, the construction yields on $G_n$ a covering chain
$x_1=x_{\mathrm{goal}},\ldots,x_{M_n}=x_{\mathrm{init}}$ of samples with
cumulative costs $C_1:=0$, $C_m:=\sum_{k=1}^{m-1}\|x_k-x_{k+1}\|$. It supplies
the following three properties, which are all that the analysis below uses.

\begin{enumerate}[label=\textup{(C\arabic*)},leftmargin=*]
	\item \textbf{Reachable steps.} $\|x_m-x_{m+1}\|\le r_n$ and the edge
	$(x_m,x_{m+1})$ is free in the current snapshot.
	\item \textbf{Clearance.} Every $x_m$ lies at distance at least $r_n$ from
	$\mathcal{X}_{\mathrm{obs}}$, so $B(x_m;r_n)$ contains no obstacle and no
	collision check on a connection to $x_m$ of length at most $r_n$ can fail
	\cite[proof of Lem.~4.2]{janson2015fast}. In D-FMT$^{*}$ every candidate is
	within $r_n$, so this applies directly to the parent search of
	Algorithm~\ref{alg:Expand}.
	\item \textbf{Length.} $C_{M_n}\le L_n$, and the covering parameters may be
	chosen so that $L_n\le(1+\zeta)c^*$ for all large $n$
	\cite[Thm.~4.1]{janson2015fast}.
\end{enumerate}

\noindent
Write $\ell_0$ for $\mathrm{lmc}$ immediately after $\mathtt{UpdateObstacles}$
and $\ell$ for $\mathrm{lmc}$ after the drain. Samples that remain in collision
are never chain vertices on $G_n$ and play no role below.

\paragraph{Repair properties.}
The analysis uses only properties \textup{(P1)}--\textup{(P3)} established in
Section~\ref{sec:algorithms} (what an update does to the labels, what it
places in $V_{\mathrm{open}}$, and how the wavefront then expands),
together with $\epsilon=0$ in Algorithm~\ref{alg:queuenode} and the full-drain
hypothesis \textup{(H2)}. The FMT\textsuperscript{X} extension additionally uses
the insertion and neighborhood-retention properties of
Section~\ref{sec:fmtx}. The practical stop rule and $\epsilon>0$ are not part
of the asymptotic-optimality result.
Two consequences of \textup{(P3)} are used throughout without comment. Labels
are monotone within a snapshot, so $\ell_0(v)\le B$ implies $\ell(v)\le B$, and
a vertex whose label is written during a drain is necessarily extracted before
that drain ends, at which point its $g$ equals its final $\mathrm{lmc}$.
Termination additionally uses positivity of edge costs, so we assume throughout
this section that every steering-feasible connection between distinct
states has strictly positive cost, as each cost functional of
Section~\ref{sec:experimental_setup} does. Self-connections never arise, since
$N^{\pm}(x)\subseteq V\setminus\{x\}$ by definition.

We now state the four lemmas that carry the argument. Their proofs are given
in \ref{app:proofs}.

\begin{lemma}[Termination]
	\label{lem:termination}
	On a fixed finite sample set, a drain satisfying \textup{(P1)} and
	\textup{(P3)} terminates.
\end{lemma}

Draining bounds the work. It says nothing by itself about the labels left
behind. The following lemma supplies that missing property, and is the only
place where the seeding rule is needed.

\begin{lemma}[Drain consistency of D-FMT$^{*}$]
	\label{lem:consistency}
	Let $\ell$ be the labels at the end of a static interval's drain. For every
	$x\in V$ and every $y\in N^{+}(x)$ whose edge is free in that interval's
	snapshot,
	\begin{equation}
		\label{eq:consistency}
		\ell(x)\;\le\;\ell(y)+d_\pi(x,y),
	\end{equation}
	unless a parent search at $x$ triggered by $y$ was resolved by a failed
	collision check in that interval or in an earlier one and the resulting
	pairwise inconsistency has not since been repaired. Either side may be
	infinite.
\end{lemma}

\noindent
Call such a pair an unresolved suboptimal connection.
This is the finite-sample suboptimal-connection event of FMT$^{*}$, in which the locally
cheapest open connection is blocked while a better parent has already left
$V_{\mathrm{open}}$. For D-FMT$^{*}$, within the interval under analysis
\textup{(C2)} already excludes it, since FMT$^{*}$ establishes that no connection
attempted at a chain vertex can fail its check. The corresponding statement for
FMT\textsuperscript{X} is established separately in
Corollary~\ref{cor:fmtx-ao}.

The chain is rebuilt from the current free space at every interval, so its
vertices always carry that clearance. What clearance cannot do is undo a label
written earlier. A vertex that is well clear now may have sat beside an obstacle
at an earlier interval, connected suboptimally there, and kept the resulting
label. Nothing
about the geometry of the current snapshot speaks to that. The reason it does
not happen is that the obstacle responsible must have departed for the vertex to
be clear now, and its departure reseeds the entire neighborhood.

\begin{lemma}[Suboptimal connections are not inherited]
	\label{lem:no-defer}
	In D-FMT$^{*}$, where every stored edge satisfies $d_\pi\le r_n$, on $G_n$ no
	consecutive chain pair carries an unresolved suboptimal connection from an
	earlier interval.
	Together with \textup{(C2)}, the exception of Lemma~\ref{lem:consistency}
	fires for no consecutive chain pair in any interval.
\end{lemma}

Consistency is what makes the covering-chain bound a telescoping sum. On
$G_n$ each chain edge is free with $\|x_m-x_{m+1}\|\le r_n$, and by
Lemma~\ref{lem:no-defer} no consecutive chain pair carries an unresolved
suboptimal connection, so
Lemma~\ref{lem:consistency} applies to every consecutive pair and, starting
from $\ell(x_1)=0$, bounds the whole chain.

\begin{lemma}[Repaired path quality]
	\label{lem:path-bound}
	For D-FMT$^{*}$, under \textup{(H1)}--\textup{(H3)} and
	\textup{(P1)}--\textup{(P3)}, $P(c_n>L_n)\le P(G_n^c)$, and in particular
	$P(c_n>L_n)\to0$.
\end{lemma}

\begin{theorem}[Static-interval AO of D-FMT$^{*}$]
	\label{thm:dfmt-ao}
	Under \textup{(H1)}--\textup{(H3)}, \textup{(P1)}--\textup{(P3)}, and the
	assumptions of Theorem~4.1 of~\cite{janson2015fast}, for every
	$\zeta>0$
	\begin{equation}
		\label{eq:dfmt-ao}
		\lim_{n\to\infty}P\bigl(c_n>(1+\zeta)\,c^*\bigr)=0,
	\end{equation}
	where $c^*$ is optimal in the current static free space.
\end{theorem}
\begin{proof}
	Lemma~\ref{lem:path-bound} gives $P(c_n>L_n)\to0$, and choosing covering
	parameters as in Theorem~4.1 of~\cite{janson2015fast} yields
	$L_n\le(1+\zeta)c^*$ for all large $n$.
\end{proof}

\begin{corollary}[Uniform over a mission]
	\label{cor:mission-ao}
	Suppose a mission contains at most $K$ replanning instants with $K$
	independent of $n$. Let $c_n^{(k)}$ be the repaired cost at instant $k$ and
	$c^{*(k)}$ the optimum in the snapshot then current. Then for every $\zeta>0$,
	$\lim_{n\to\infty}P\bigl(\exists\,k\le K:\;c_n^{(k)}>(1+\zeta)c^{*(k)}\bigr)=0$.
\end{corollary}
\begin{proof}
	Theorem~\ref{thm:dfmt-ao} applies at each fixed $k$, and a union bound over
	the $K$ instants preserves the limit.
\end{proof}

\begin{remark}
	We do not claim equal trees or $c_n\le c_n^{\mathrm{FMT}^*}$. AO uses only
	$c_n\le L_n$ on $G_n$. Nodes remaining in collision may stay at infinite cost
	and never appear on the free covering chain. The exception clause of
	Lemma~\ref{lem:consistency} is FMT$^{*}$'s own suboptimal-connection event
	and is not introduced by repair. The parent-to-child push of Phase~1 addresses
	a different failure, namely stale descendants beneath an already-correct
	parent, and is not claimed to remove it. Neither \textup{(C2)} nor
	Lemma~\ref{lem:no-defer} removes it. They show only that it cannot affect the
	consecutive chain pairs used by the covering-chain bound, which is all that
	bound requires. Convergence is in probability, as
	in~\cite{janson2015fast}.
\end{remark}

\begin{corollary}[Asymptotic optimality of FMT\textsuperscript{X}]
	\label{cor:fmtx-ao}
	Under the standard geometric and sampling assumptions of the
	RRT$^{*}$/RRT\textsuperscript{X} asymptotic-optimality result, and relying on
	the asymptotically optimal chain property of the corresponding sequential
	sampling, admission, and retained-neighborhood construction, geometric
	FMT\textsuperscript{X} with $\epsilon=0$ in $\mathtt{QueueNode}$,
	full-drain $\mathtt{Expand}$ on each static interval, and all birth
	$r_n$-neighborhood links retained, satisfies~\eqref{eq:dfmt-ao}.
\end{corollary}
\begin{proof}
	The sampling, saturation, admission, and retained
	birth-neigh\-bor\-hood construction of $\mathtt{Extend}$ follows the RRT\textsuperscript{X} graph
	construction~\cite[Alg.~2]{otte2016rrtx}. We rely on the corresponding
	RRT$^{*}$/RRT\textsuperscript{X} graph result for the existence of an
	asymptotically optimal chain in the retained candidate graph. As in the
	standard RRT$^{*}$ asymptotic-optimality construction, this chain is obtained
	along a sufficiently clear approximating path, so the construction provides an
	obstacle-free local neighborhood about each chain vertex containing the
	corresponding chain connection~\cite{karaman2011sampling}. It remains to show that
	FMT\textsuperscript{X}'s lazy parent selection and wavefront repair do not
	destroy that cost bound.

	Although FMT\textsuperscript{X} neighborhoods may become asymmetric under
	culling, consecutive chain connections have length at most the current $r_n$.
	Since $r_n$ is nonincreasing, such a connection has never exceeded the active
	radius, so its reciprocal record has not been culled. Extraction of a chain
	predecessor $z$ therefore reaches its successor $x$ through the
	reverse-neighborhood sweep.

	Consider a chain relaxation of $x$ from an extracted predecessor $z$.
	Because $z$ has minimum $\mathrm{lmc}$ in $V_{\mathrm{open}}$ and remains open
	during its neighbor sweep, $z$ is itself a candidate, so the selected parent
	satisfies
	$\mathrm{lmc}(y_{\min})+d_\pi(x,y_{\min})\le\mathrm{lmc}(z)+d_\pi(x,z)$ while
	$\mathrm{lmc}(y_{\min})\ge\mathrm{lmc}(z)$, whence
	\[
	d_\pi(x,y_{\min})\;\le\;d_\pi(x,z).
	\]
	The selected connection is therefore no longer than the corresponding chain
	connection and remains within the same obstacle-free local neighborhood, so
	its collision check succeeds. Lazy parent selection can
	thus never replace a valid chain connection by a longer competing one, and
	retained birth links exceeding the current $r_n$ can never be the connection
	that fails a check when the chain relaxation is performed.

	The same bound covers an inherited failure that could have prevented
	consistency of this chain pair. Whenever $z$ was the triggering open
	candidate, any selected candidate that failed its check had
	$d_\pi(x,y_{\min})\le d_\pi(x,z)$,
	so its blocking obstacle lay inside the local neighborhood that the current
	chain requires to be obstacle-free. If that obstruction is absent, the
	corresponding obstacle change activates \textup{(P2)}, which reseeds the
	finite-$\mathrm{lmc}$ neighbors needed to restore the relaxation under a full
	drain.
	Finally, the FMT\textsuperscript{X} repair satisfies
	\textup{(P1)}--\textup{(P3)} as stated in Section~\ref{sec:algorithms}, with
	$\rho=\delta$ in \textup{(P2)}, since $\mathtt{Extend}$ inserts every surviving
	sample into $V_{\mathrm{open}}$ and the drain extracts it.
	The preceding argument therefore establishes the pairwise inequalities
	required along the chain, and the same telescoping bound as in
	Lemma~\ref{lem:path-bound} yields~\eqref{eq:dfmt-ao}.
\end{proof}

\begin{remark}
	\label{rem:positive-eps}
	Corollary~\ref{cor:fmtx-ao} is stated for $\epsilon=0$, while the executed
	planner uses the anytime cascade filter $\epsilon>0$
	of~\cite{otte2016rrtx}. This does not leave a suppressed vertex stranded:
	each later $\mathtt{Extend}$ inserts its new samples into
	$V_{\mathrm{open}}$, and extracting one relaxes exactly the neighbors whose
	cascades were cut short, with their children reached by the parent-to-child
	push. Under continued densification such vertices are therefore revisited
	rather than frozen. We do not prove this, and state the guarantee at
	$\epsilon=0$.
\end{remark}

\section{Complexity Analysis}
\label{sec:complexity_analysis_summary}

In this section, we analyze the computational complexity of the FMT\textsuperscript{X} algorithm,
focusing on its anytime state‑space expansion and dynamic replanning capabilities.
Our analysis relies on the following standard assumptions for a spatial graph of $n$ samples:
\begin{itemize}
	\item Priority queue ($V_{\mathrm{open}}$) operations take $\mathcal{O}(\log n)$ time.
	\item A single collision check (potentially involving complex spatial or temporal continuous
	collision detection) costs $T_{\mathrm{coll}}$.
	\item Steering between two states costs $T_{\mathrm{steer}}$.
	\item A spatial index supports $r_n$-ball queries in $\mathcal{O}(\log n)$ time.
	Under the asymptotically optimal radius scaling $r_n$~\cite{janson2015fast} and
	the shrinking-radius and retained-birth-neighborhood culling construction of
	RRT\textsuperscript{X}~\cite{otte2016rrtx}, the maintained neighborhood size
	satisfies $\Delta_n = \mathcal{O}(\log n)$ in expectation.
\end{itemize}
We denote the following aggregated quantities per replanning cycle (time step $t_k \to t_{k+1}$):
\begin{itemize}
	\item $m$ -- number of new samples added during the anytime expansion phase,
	\item $k$ -- number of nodes extracted from $V_{\mathrm{open}}$ during the wavefront expansion,
	\item $N_{\mathrm{aff}}$ -- total number of candidate tree nodes potentially invalidated across
	all obstacle additions,
	\item $N_{\mathrm{orphan}}$ -- total number of tree nodes actually orphaned across all obstacle additions,
	\item $N_C$ -- total number of unique boundary neighbors queued as a result of all obstacle additions,
	\item $N_{\mathrm{rec}}$ -- total number of unique nodes returned by the recovery spatial queries
	across all obstacle removals.
\end{itemize}

\subsection{Incremental State‑Space Expansion}
For each newly sampled state in a batch (Algorithm~\ref{alg:Extend}), the expected cost of the
following operations is:
\begin{enumerate}
	\item Query the spatial index for neighbors: $\mathcal{O}(\log n)$ time, retrieving
	$\Delta_n = \mathcal{O}(\log n)$ candidates in expectation.
	\item For each candidate, execute the local steering evaluations needed to build the
	stored neighbor links:
	$\mathcal{O}(\Delta_n \cdot T_{\mathrm{steer}})$ in expectation.
	\item Evaluate candidates in increasing order of estimated cost
	$\mathrm{lmc}(y) + d_\pi(x,y)$. By the obstacle-shell volume argument underlying
	Lemma~C.2 of~\cite{janson2015fast}, the expected number of candidate edges
	collision‑checked per new sample is $\mathcal{O}(1)$. Hence the expected
	collision‑checking cost of the parent-selection step is $\mathcal{O}(T_{\mathrm{coll}})$.
	\item If a valid parent is found, insert the node into $V_{\mathrm{open}}$: $\mathcal{O}(\log n)$ expected.
\end{enumerate}
Summing over all $m$ samples in the batch, the expected time for the incremental expansion phase is
\begin{equation}
	\mathcal{O}\bigl( m\log n \;+\; m\,\Delta_n\,T_{\mathrm{steer}} \;+\; m\,T_{\mathrm{coll}} \bigr).
\end{equation}
Because the batch size $m$ is constant, this simplifies to $\mathcal{O}(\log n + \Delta_n T_{\mathrm{steer}}
+ T_{\mathrm{coll}})$ per cycle.

Over the lifetime of the algorithm, as $n$ nodes are added, the expected total
number of collision checks across all batches is $\mathcal{O}(n)$. Samples that
land far from obstacles are connected after a single successful collision check,
which contributes $\mathcal{O}(n)$ checks. A sample inserted at graph size $j$
can require additional checks only when an obstacle lies within its birth radius
$r_j$, and such a sample is checked at most as often as it has neighbors,
$\mathcal{O}(\log n)$ in expectation. By the obstacle-shell volume argument
underlying Lemma~C.2 of~\cite{janson2015fast}, these extra checks total
\(\mathcal{O}\bigl(\log n\sum_{j=2}^{n} r_j\bigr)
=\mathcal{O}(n\,r_n\log n)=o(n)\),
so the expected lifetime total remains $\mathcal{O}(n)$.

\subsection{Dynamic Graph Repair}
During the $\mathtt{UpdateObstacles}$ phase, the algorithm processes all environmental changes that occurred
since the previous cycle.  For each obstacle addition (Algorithm~\ref{alg:AddObstacle}), a spatial query
retrieves an aggregated candidate set of size $N_{\mathrm{aff}}$.  Each candidate edge is checked against the
new obstacle, costing $\mathcal{O}(N_{\mathrm{aff}}\,T_{\mathrm{coll}})$.  Nodes whose parent edges intersect
an obstacle, together with all of their tree descendants, form the orphan set
$\mathcal{V}_{\mathcal{T}}^{\mathrm{orphan}}$ of size $N_{\mathrm{orphan}}$, which may therefore be
larger than $N_{\mathrm{aff}}$.
For each orphan, removal from $V_{\mathrm{open}}$ (if present) and cost resetting take $\mathcal{O}(N_{\mathrm{orphan}}\log n)$.
To seed the repair wavefront, the algorithm iterates over all forward neighbors $y\in N^{+}(\mathcal{V}_{\mathcal{T}}^{\mathrm{orphan}})$
($\mathcal{O}(N_{\mathrm{orphan}}\Delta_n)$ time) and inserts those with finite $\mathrm{lmc}$ that are not
already in $V_{\mathrm{open}}$ ($\mathcal{O}(N_C\log n)$ time).

For each obstacle removal (Algorithm~\ref{alg:RemoveObstacle}), the recovery spatial query
returns a conservative set of candidate nodes $\mathcal{V}^{\mathrm{rec}}$ of size
$N_{\mathrm{rec}}$. Each candidate node with finite $\mathrm{lmc}$ is inserted into
$V_{\mathrm{open}}$ if not already present, costing
$\mathcal{O}(N_{\mathrm{rec}}\log n)$.

Because $\Delta_n = \mathcal{O}(\log n)$, the neighbor-scan term for the
orphan set merges with its queue term. Hence the total worst-case
complexity for processing a batch of dynamic obstacle changes within a single
replanning cycle is
\begin{equation}
	\mathcal{O}\bigl( N_{\mathrm{aff}}\,T_{\mathrm{coll}} \;+\; (N_{\mathrm{orphan}} + N_C + N_{\mathrm{rec}})\log n \bigr).
\end{equation}
In a global worst‑case update that affects the entire state space simultaneously, each of these quantities
can be as large as $\mathcal{O}(n)$, leading to a per‑cycle repair cost of $\mathcal{O}(n\,T_{\mathrm{coll}} + n\log n)$.

\subsection{Wavefront Expansion and Suboptimality Repair}
\label{sec:wavefront-expansion}

Let $k$ be the number of nodes extracted from $V_{\mathrm{open}}$ during a single wavefront expansion
(repair cycle).  The expansion runs on a static obstacle snapshot, as no obstacle changes occur while
the wavefront is active, so the graph $G$ and its collision‑free edges are fixed for the duration of
the cycle.
For each extracted node $z$ the following operations are performed:
\begin{itemize}
	\item Extract $z$ from the priority queue: $\mathcal{O}(\log n)$.
	\item Propagate $z$'s new cost to its tree children. The tree-child relation is
	maintained separately from the neighbor sets, so the culling bound below does not
	apply to it. Instead, since the shortest-path tree contains at most $n-1$
	parent-child edges, the total child-propagation work over a complete one-pass
	traversal is $\mathcal{O}(n)$.
	\item \texttt{CullNeighbors}$(z)$: iterate over $z$'s directed neighbors and remove edges
	that are now longer than the connection radius, in $\mathcal{O}(\Delta_n)$ time.
	\item For each neighbor $x\in N^{-}(z)$:
	\begin{itemize}
		\item The cost-improvement condition $\mathrm{lmc}(x) > g(z) + d_\pi(x,z)$ is checked
		in $\mathcal{O}(1)$ time.
		\item If the condition holds ($x$ is suboptimal), the algorithm performs a
		Bellman‑like parent search, which includes:
		\begin{itemize}
			\item \texttt{CullNeighbors}$(x)$ ($\mathcal{O}(\Delta_n)$),
			\item scanning the candidate parent set $Y_{\mathrm{better}}
			\subseteq N^{+}(x)\cap V_{\mathrm{open}}$ ($\mathcal{O}(\Delta_n)$),
			\item one collision check if the best candidate differs from the current parent
			($T_{\mathrm{coll}}$), and
			\item updating or inserting $x$ in the priority queue ($\mathcal{O}(\log n)$).
		\end{itemize}
	\end{itemize}
\end{itemize}

Let $R$ be the number of executions of the full parent-search block during the cycle,
counting repeated executions at the same vertex separately, as in the node-consideration
count of~\cite[Lem.~C.2]{janson2015fast}.  Each such occurrence costs
$\mathcal{O}(\Delta_n + \log n + T_{\mathrm{coll}}) = \mathcal{O}(\log n + T_{\mathrm{coll}})$.
The total time of the wavefront expansion is therefore
\begin{equation}
	\mathcal{O}\bigl( k \log n \;+\; k \Delta_n \;+\; |R| (\log n + T_{\mathrm{coll}}) \bigr).
\end{equation}
Because $\Delta_n = \mathcal{O}(\log n)$, the term $k \Delta_n$ is absorbed by $k \log n$,
and the $\mathcal{O}(n)$ child-propagation work is absorbed by the cumulative snapshot
bound below.

\medskip
\noindent\textbf{Bounding \(|R|\) in one static snapshot.}
Consider a single wavefront expansion on a fixed obstacle configuration. Each vertex
contributes at most one first consideration in the current snapshot, including vertices
reconsidered as part of obstacle-triggered repair, accounting for at most \(n\) events.
Away from a failed lazy collision check, the cost-ordered parent search resolves the
vertex without creating another source of reconsideration. By
Lemma~\ref{lem:cost-equivalence} and the argument of
Section~\ref{subsec:goal-rooted-interpretation}, the strict cost-improvement predicate
does not trigger another parent search at that vertex during the same snapshot unless
lazy obstacle handling has left it with a suboptimal finite label. Any additional
consideration must therefore follow the same obstacle-local failed-check mechanism.
For D-FMT$^{*}$ every stored edge has length at most the fixed \(r_n\), so such repeated
considerations are confined to vertices within \(r_n\) of an obstacle. Such a vertex is
considered at most as often as it has neighbors, \(O(\log n)\) in expectation, while the
expected number of vertices within \(r_n\) of an obstacle is \(O(n\,r_n)\). The
obstacle-shell counting argument of Lemma~C.2 of~\cite{janson2015fast} therefore gives
\(O(n\,r_n\log n)=O\bigl(n^{1-1/d}(\log n)^{1+1/d}\bigr)=o(n)\) expected repeats.
FMT\textsuperscript{X} additionally retains birth edges. A vertex inserted at graph size
\(j\) may therefore retain edges of length up to \(r_j\). Replacing the uniform
\(n\,r_n\) shell count of Lemma~C.2 by the corresponding birth-radius count
\(\sum_{j\le n} r_j = O(n\,r_n)\) yields the same \(o(n)\) expected number of
repeated considerations. The stated identity follows from
\[
\begin{aligned}
	\sum_{j=2}^{n} r_j
	&\;\le\; C\gamma(\log n)^{1/d}\sum_{j=2}^{n} j^{-1/d}\\
	&\;=\; O\bigl(n^{1-1/d}(\log n)^{1/d}\bigr)
	\;=\; O(n\,r_n).
\end{aligned}
\]
Hence the expected total number of parent-search executions \(|R|\) in the snapshot
is \(O(n)\).
Combined with the per‑extraction time, the cumulative work of the snapshot is
\(O(n\log n + n\,T_{\mathrm{coll}})\) and the number of collision checks is \(O(n)\), matching the asymptotic efficiency of static FMT$^{*}$. If obstacles change more frequently or
affect larger portions of the tree, the per-cycle formulas above still hold and can be summed
according to the specifics of the application.

\begin{figure*}[!t]
	\centering
	
	\begin{subfigure}[b]{0.33\textwidth}
		\includegraphics[width=\textwidth]{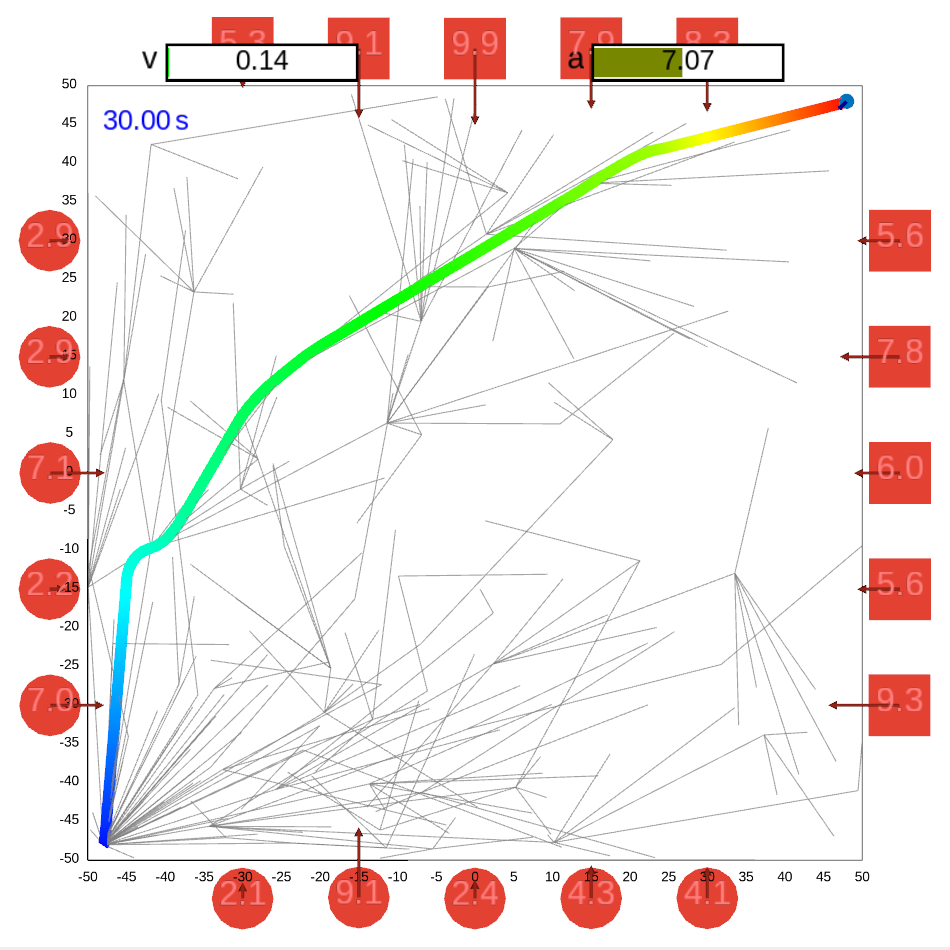}
	\end{subfigure}\hfill
	\begin{subfigure}[b]{0.33\textwidth}
		\includegraphics[width=\textwidth]{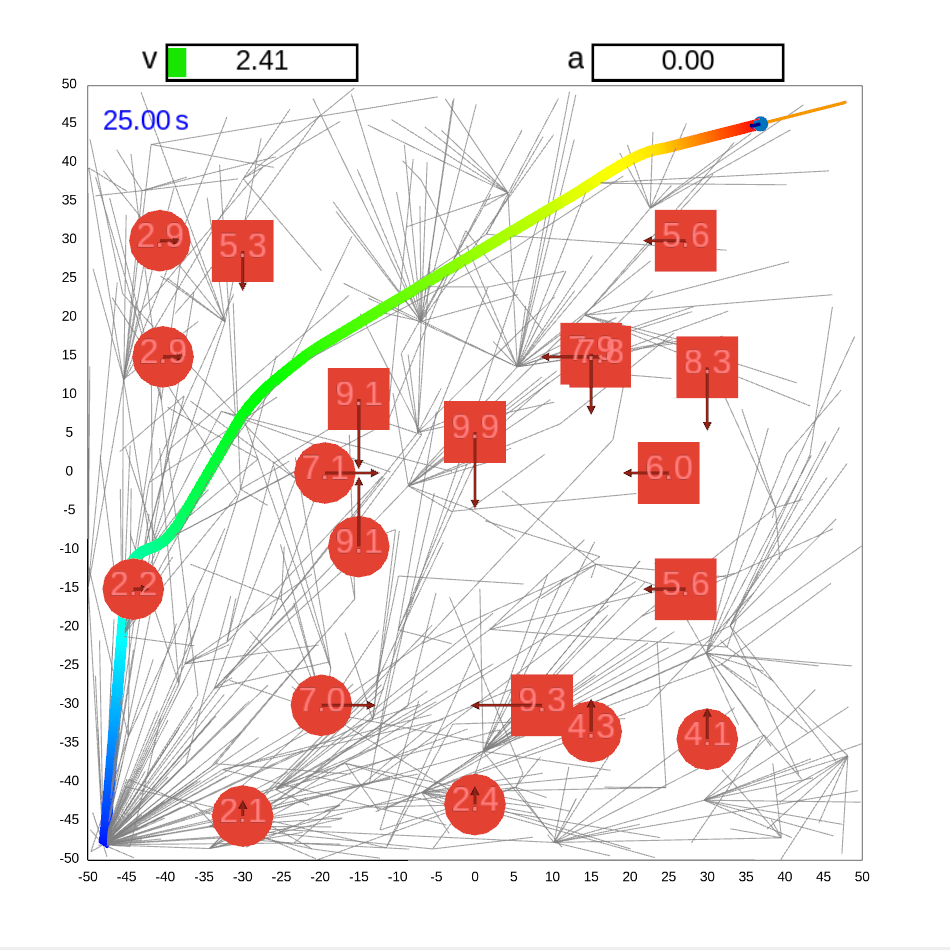}
	\end{subfigure}\hfill
	\begin{subfigure}[b]{0.33\textwidth}
		\includegraphics[width=\textwidth]{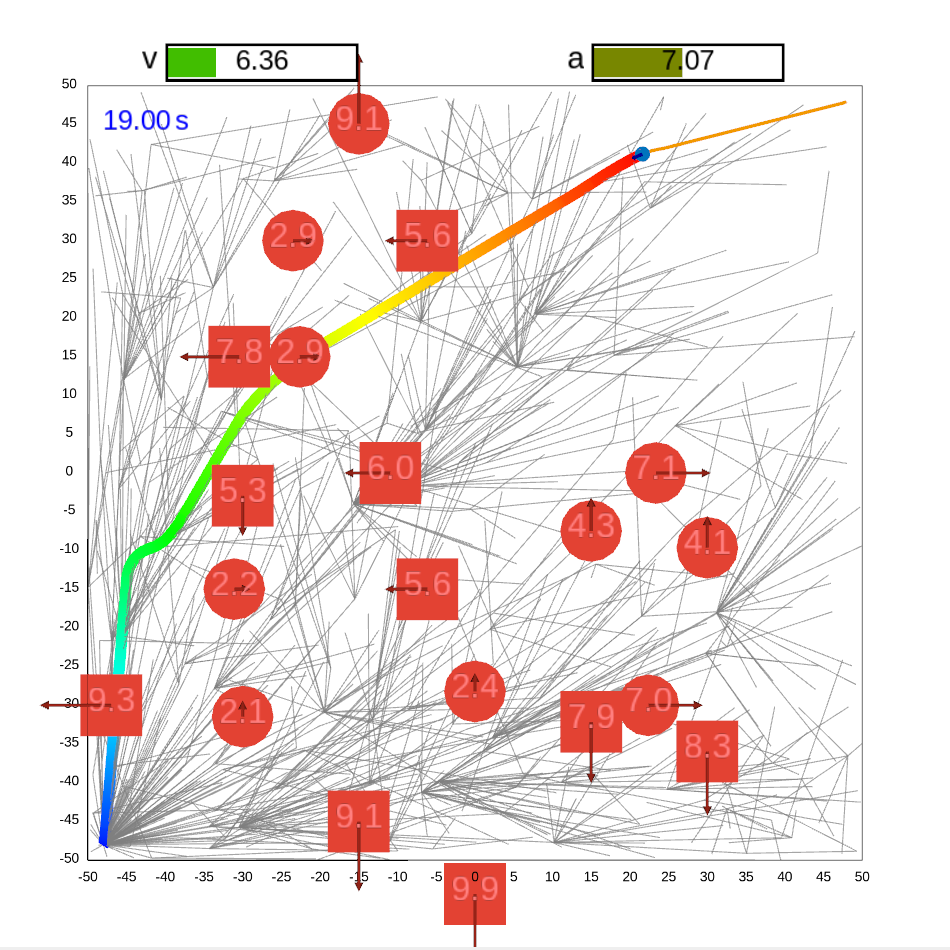}
	\end{subfigure}
	\\[4pt] 
	
	\begin{subfigure}[b]{0.33\textwidth}
		\includegraphics[width=\textwidth]{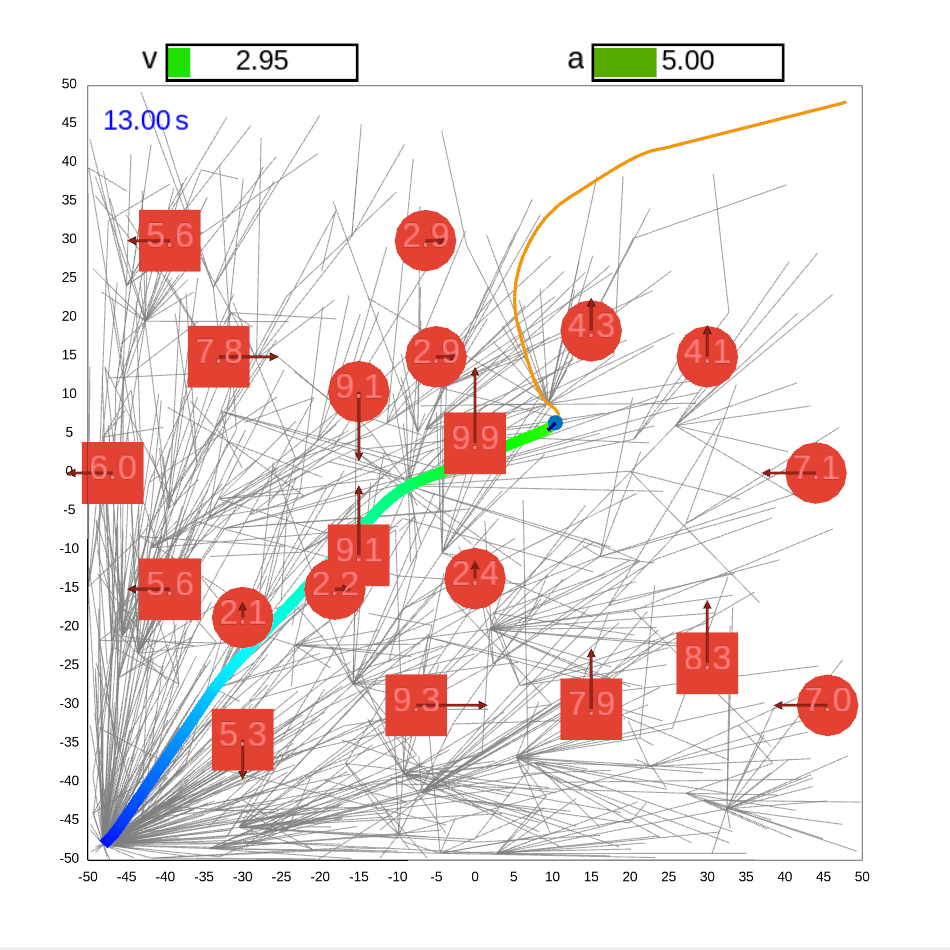}
	\end{subfigure}\hfill
	\begin{subfigure}[b]{0.33\textwidth}
		\includegraphics[width=\textwidth]{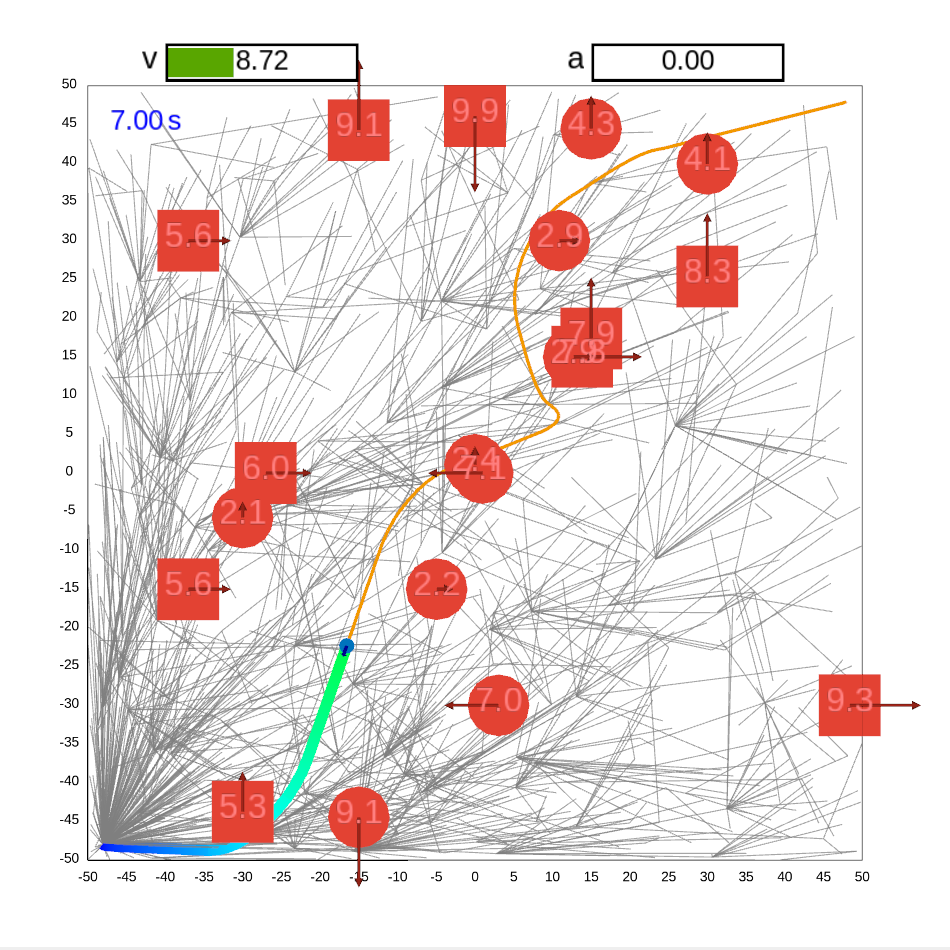}
	\end{subfigure}\hfill
	\begin{subfigure}[b]{0.33\textwidth}
		\includegraphics[width=\textwidth]{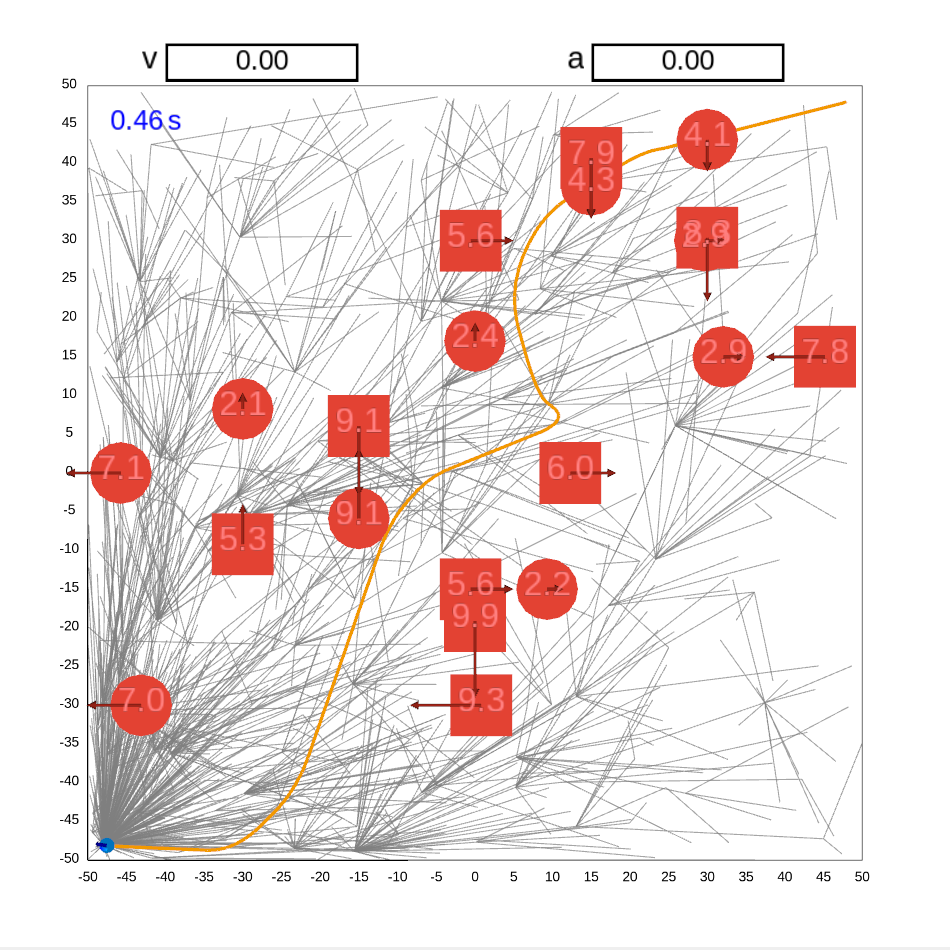}
	\end{subfigure}
	\\[4pt] 
	
	\caption{A sequential visualization of the FMT\textsuperscript{X} replanning
	process for the second-order thruster model as the robot and obstacles move
	through the environment. The numbers inside the obstacles denote their
	velocity magnitudes, and the arrows indicate their velocity directions.}
	\label{fig:thruster_evolution}
\end{figure*}

\section{Experiments}

We evaluate the proposed replanners through comparative benchmarks in dynamic
environments. D-FMT$^{*}$ is compared with D$^{*}$~Lite on identical
fixed PRM$^{*}$ graphs, while FMT\textsuperscript{X} is compared with
RRT\textsuperscript{X} and LLPT$^{*}$ under online densification. We begin in
geometric $\mathbb{R}^{2}$ to isolate graph-repair performance and then
evaluate the planners in kinodynamic state spaces.

\subsection{Simulation Environment and Evaluation Protocol}

We simulate moving obstacles whose trajectories are not known a priori by the
planners and do not react to the robot. To isolate replanning from perception
uncertainty, obstacle observations are supplied by a deterministic oracle. At
replanning time $t_k$, the oracle provides the exact obstacle position
$p_{\mathrm{obs}}(t_k)$, geometry, and velocity $v_{\mathrm{obs}}(t_k)$.
Writing $\tau\ge0$ for look-ahead time from the replanning instant, future
motion is predicted linearly as
\[
\hat{p}_{\mathrm{obs}}(t_k+\tau)
=
p_{\mathrm{obs}}(t_k)+v_{\mathrm{obs}}(t_k)\tau .
\]
In the fully observable experiments, a dynamic replanning event
$\Delta\mathcal{X}_{\mathrm{obs}}\neq\emptyset$ is generated when an
obstacle changes its trajectory, such as at a waypoint. The planners then
invalidate affected regions and repair using the updated linear prediction.
In the partial-observability study, newly revealed obstacles also generate
updates.

All planners are evaluated in a synchronous simulation with slice duration
$\Delta t=0.02\,\mathrm{s}$. At each cycle the planner receives a nominal
computational slice of $\Delta t$, during which it processes obstacle updates
and, for the anytime planners, sample addition. The simulation clock then
advances by exactly $\Delta t$, while computation time is recorded separately.
All planners therefore experience identical obstacle trajectories in simulated
time.

All experiments were performed on an
Intel\textsuperscript{\textregistered}
Core\textsuperscript{\texttrademark} i7-4750HQ at $2.00\,\mathrm{GHz}$ with
$16\,\mathrm{GB}$ DDR3 memory running Ubuntu 24.04 LTS.\footnote{Source code
	and reproduction instructions are available at
	\url{https://github.com/sohail70/motion_planning}.}

\subsubsection{Statistical Aggregation and Metrics}
\label{sec:experiments_protocol_metrics}

Graph topology varies with the random seed used for sampling. We therefore
evaluate 100 seeds per experimental condition and compare planners on identical
seeds. Latency and quality statistics use only common-success seeds, meaning
seeds on which every compared planner satisfies the success criterion below.
This pairing controls instance variation but conditions the reported latency
and quality statistics on jointly solvable instances. Success rates are
therefore reported separately over all 100 seeds.

We report the following metrics.
\begin{itemize}
	\item \textbf{Repair latency.}
	For each seed, we compute the median repair time over all repair events.
	Reported values are the median and interquartile range of these per-seed
	medians.
	
	\item \textbf{Solution quality.}
	We report executed path length, traversal time, total heading change for
	Dubins, and control effort $\int |a|\,dt$ for Thruster, where $|a|$ denotes
	acceleration magnitude. Control effort is accumulated numerically along the
	executed trajectory as $\sum_i\|v_{i+1}-v_i\|_2$.
	
	\item \textbf{Collision checks.}
	We count the collision checks associated with each obstacle-triggered repair.
	For each seed, the mean number of checks per repair is computed first, followed
	by the median and interquartile range across paired seeds.

	\item \textbf{Success rate.}
	The fraction of all 100 seeds in which the robot reaches the goal while
	maintaining a collision-free trajectory throughout execution.
	
	\item \textbf{Observed maximum latency.}
	The largest individual repair latency over the paired common-success seeds,
	used to characterize the observed extreme tail alongside the median and
	interquartile range.
\end{itemize}

\subsection{Experimental Setup}
\label{sec:experimental_setup}

All experiments use a $100\,\mathrm{m}\times100\,\mathrm{m}$ workspace, a
goal-region radius of $0.5\,\mathrm{m}$, and an obstacle-inflation radius of
$1.0\,\mathrm{m}$ accounting for the robot radius and safety margin. The
connection radius is
\[
r_n=
\min\left\{
\delta,\;
C\gamma\left(\frac{\log n}{n}\right)^{1/d}
\right\},
\]
where $\gamma$ follows the theoretical lower bound, $C$ is a scaling factor,
and $\delta$ is the steering-extension bound.

\subsubsection{State Spaces and Cost Functionals}

We evaluate four state spaces. For a trajectory $\pi$ composed of edges between
states $x_i$ and $x_{i+1}$, the total cost is
\[
J(\pi)=\sum_i d_\pi(x_i,x_{i+1}),
\]
with edge cost defined according to the state space.

\begin{itemize}
	\item \textbf{Geometric ($\mathbb{R}^2$).}
	Straight-line Euclidean distance
	\[
	d_\pi(x,y)
	=
	\|p_y-p_x\|_2 .
	\]
	
	\item \textbf{Holonomic ($\mathbb{R}^2\times\mathbb{T}$).}
	Weighted Euclidean distance with the forward-time constraint $t_y>t_x$
	\[
	d_\pi(x,y)
	=
	\sqrt{
		c_t(t_y-t_x)^2
		+
		\|p_y-p_x\|_2^2
	},
	\qquad c_t>0.
	\]
	
	\item \textbf{Dubins
		($\mathbb{R}^2\times\mathbb{S}^1\times\mathbb{T}$).}
	The shortest Dubins curve between the endpoint configurations is combined
	with the elapsed time to define
	\[
	d_\pi(x,y)
	=
	\sqrt{
		\bigl(L_{\mathrm{Dubins}}(x,y;r_{\min})\bigr)^2
		+
		c_t^2(\Delta t_{xy})^2
	},
	\qquad c_t>0,
	\]
	where $L_{\mathrm{Dubins}}(x,y;r_{\min})$ is the shortest Dubins path length
	for minimum turning radius $r_{\min}$. Valid edges satisfy
	$t_y>t_x$ and
	\[
	v_{\min}
	\le
	\frac{L_{\mathrm{Dubins}}(x,y;r_{\min})}{\Delta t_{xy}}
	\le
	v_{\max}.
	\]
	
	\item \textbf{Thruster
		($\mathbb{R}^2_x\times\mathbb{R}^2_{\dot{x}}\times\mathbb{T}$).}
	The cost of a feasible trajectory obtained from the two-point boundary-value
	problem is
	\[
	d_\pi(x,y)
	=
	\int_{t_x}^{t_y}
	\sqrt{
		c_p^2\|\dot{p}(\tau)\|_2^2
		+
		c_v^2\|\dot{v}(\tau)\|_2^2
		+
		c_t^2
	}\,d\tau ,
	\]
	with $c_p,c_v,c_t>0$.
\end{itemize}

Obstacle speeds are sampled uniformly from $15$--$25\,\mathrm{m/s}$ in the
holonomic environment, $10$--$20\,\mathrm{m/s}$ in Dubins, and
$5$--$10\,\mathrm{m/s}$ in Thruster. Robot constraints are
$v_{\max}=20\,\mathrm{m/s}$ for the holonomic model,
$v\in[2,15]\,\mathrm{m/s}$ and $r_{\min}=2.0\,\mathrm{m}$ for Dubins, and
$v_{\max}=10\,\mathrm{m/s}$ and $a_{\max}=5\,\mathrm{m/s}^2$ for Thruster.
The geometric experiments isolate graph repair and impose no dynamic or
kinematic constraints.

\subsubsection{Fixed-Graph and Anytime Replanning Benchmarks}
\label{subsubsec:replanning_benchmarks}

We consider two complementary replanning settings. In the fixed-graph setting,
D-FMT$^{*}$ and D$^{*}$~Lite operate on the same precomputed PRM$^{*}$ graph.
D-FMT$^{*}$ handles obstacle updates through orphaning and cost-ordered repair,
while D$^{*}$~Lite applies edge-cost updates followed by incremental
shortest-path repair with an admissible heuristic. In the anytime setting,
FMT\textsuperscript{X}, RRT\textsuperscript{X}, and LLPT$^{*}$ incrementally
add samples during execution. The batch size $m$ denotes the number of samples
added per sample-addition call.

FMT\textsuperscript{X} uses the same sample-admission condition as
RRT\textsuperscript{X}, retaining a sample only when a collision-free parent of
finite cost exists within $r_n$. This is an experimental choice rather than an
algorithmic requirement. The FMT$^{*}$-native alternative is to retain unattached samples and allow
the wavefront to connect them later. Using the former rule makes reported $|V|$ represent the same
quantity for every planner, so comparisons use matched connected-vertex counts
rather than matched raw sample counts. In kinodynamic spaces it also avoids
spending the graph budget on sampled states that do not admit a feasible parent
at insertion.

For the fixed-graph benchmark, we evaluate
\[
(n,|O|)
\in
\{(1000,10),(1000,20),(3000,20)\}.
\]
The 1000-node PRM$^{*}$ graph is evaluated with 10 and 20 obstacles, while the
3000-node graph is evaluated with 20 obstacles. The primary anytime benchmark
uses
\[
(m,|O|)
\in
\{(1,10),(1,20),(3,20)\},
\]
with 500 initial samples in every run. Thus, $m=1$ is evaluated with 10 and
20 obstacles, while $m=3$ is evaluated with 20 obstacles.

Two additional anytime studies are performed. The first evaluates
time-to-goal reachability filtering in Thruster for
FMT\textsuperscript{X}, RRT\textsuperscript{X}, and LLPT$^{*}$ using
\[
T\in\{35,45,55,65\}\,\mathrm{s}.
\]
When filtering is enabled, the planner drops a node $x$ whenever
\[
\tau(x)>T_{\mathrm{robot}},
\]
where $T_{\mathrm{robot}}$ is the robot's remaining time budget. This is an
exact reachability filter rather than a heuristic pruning rule. Feasible edges
strictly decrease the time-to-goal coordinate toward the goal, while
$T_{\mathrm{robot}}$ is monotonically non-increasing during execution.
Filtered nodes therefore cannot participate in a trajectory that remains
executable by the robot.

The second study evaluates partial obstacle observability in Dubins. Obstacles
become observable only within a $20\,\mathrm{m}$ sensing range. The experiment
uses a $100\,\mathrm{s}$ time budget with reachability filtering enabled. In a
time-augmented state space, a single goal state also fixes a single arrival
time. We therefore use a multi-horizon goal set containing 100 copies of the
goal configuration that differ only in their time-to-goal coordinate. The copies correspond to evenly spaced travel times spanning the feasible
interval from the minimum travel time at maximum velocity to the full time
budget. Each copy is a zero-cost root of the
backward search, allowing the robot to terminate at an achievable arrival time
rather than waiting to match a prescribed one.

Across both primary benchmark settings, the radius scaling factor is fixed at
$C=3.0$ for every state space and obstacle configuration. The geometric
fixed-graph and anytime experiments use 50 and 500 simulation iterations,
respectively. The primary dynamic benchmarks use time budgets of
$25\,\mathrm{s}$ for holonomic, $30\,\mathrm{s}$ for Dubins, and
$35\,\mathrm{s}$ for Thruster in both replanning settings. For the anytime
planners, the inconsistency tolerance is $\epsilon=0.1$ for
FMT\textsuperscript{X} and RRT\textsuperscript{X}. Steering-extension bounds
are $\delta=10.0$, $25.0$, $30.0$, and $45.0$ for geometric, holonomic,
Dubins, and Thruster, respectively.

\subsection{Experimental Results}
\label{sec:experiments}

\subsubsection{Fixed-Graph Replanning}

\begin{figure*}[!t]
	\centering
	\includegraphics[width=\textwidth]{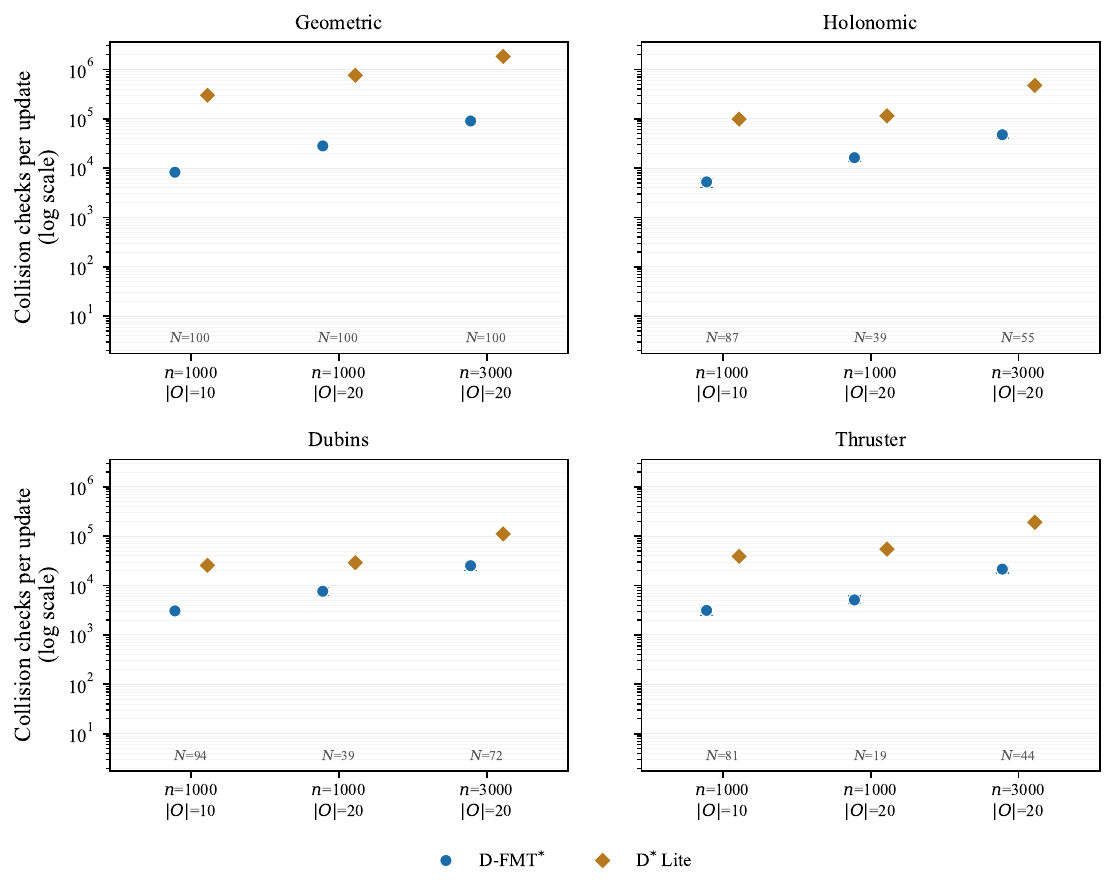}
	\caption{Fixed-graph benchmark. Collision checks per update for
		D-FMT$^{*}$ and D$^{*}$~Lite on a logarithmic scale. $N$ denotes the
		number of paired common-success seeds for each configuration.}
	\label{fig:edge_checks_fixed}
\end{figure*}

\begin{figure*}[!t]
	\centering
	\includegraphics[width=\textwidth]{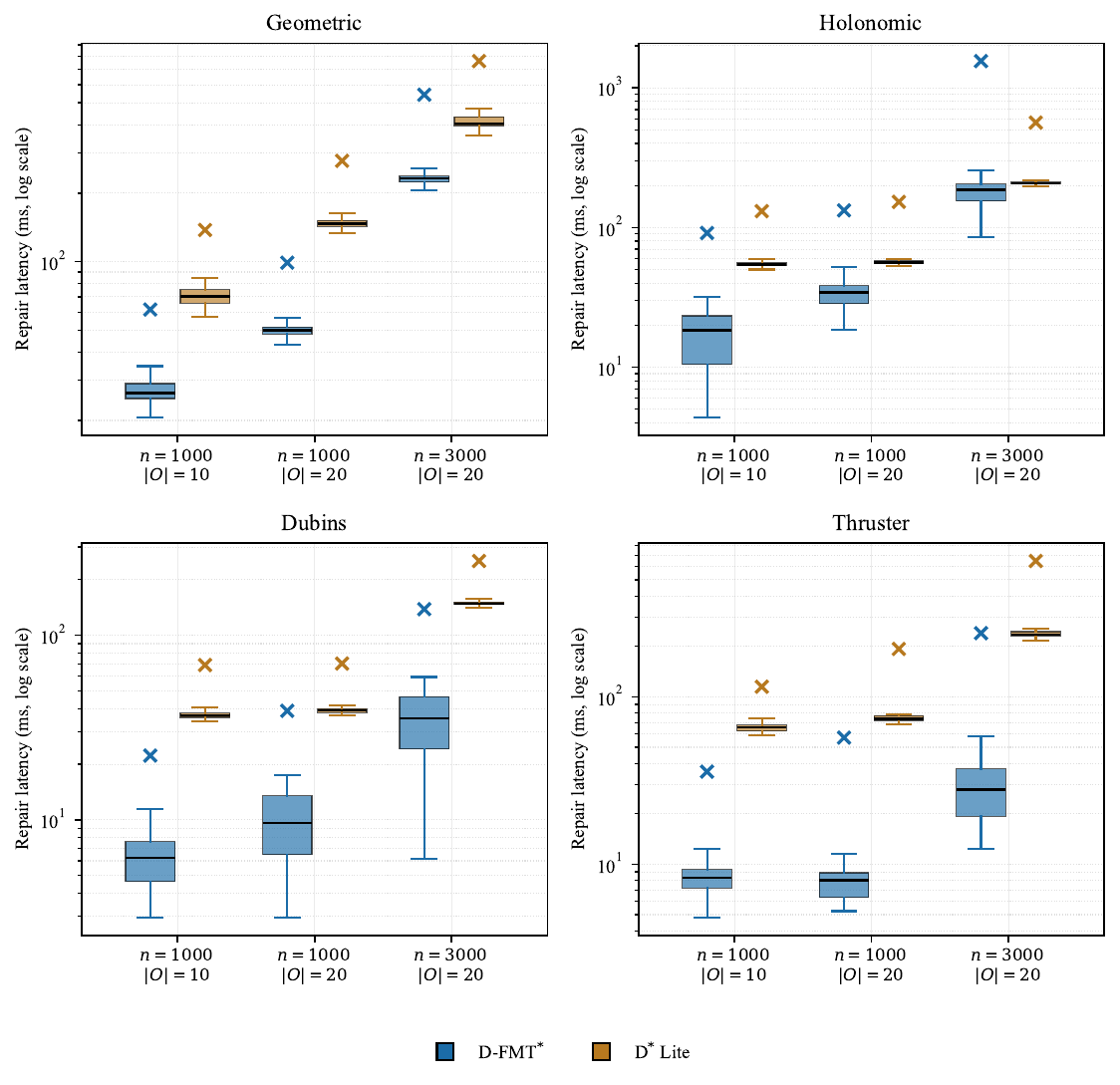}
	\caption{Fixed-graph benchmark. Repair latency of D-FMT$^{*}$ and
		D$^{*}$~Lite across graph size $n$ and obstacle count $|O|$.}
	\label{fig:na_repair_time}
\end{figure*}

\begin{figure*}[!t]
	\centering
	\includegraphics[width=\textwidth]{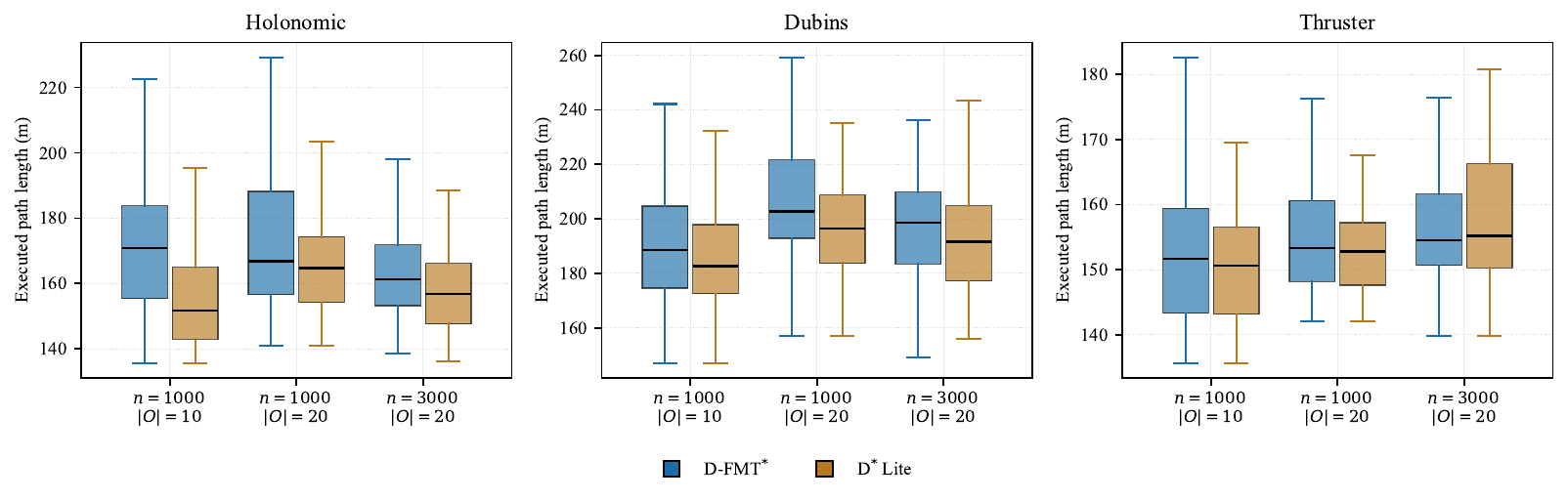}
	\caption{Fixed-graph benchmark. Executed path length across graph size
		$n$ and obstacle count $|O|$.}
	\label{fig:na_path_length}
\end{figure*}

\begin{figure*}[!t]
	\centering
	\includegraphics[width=\textwidth]{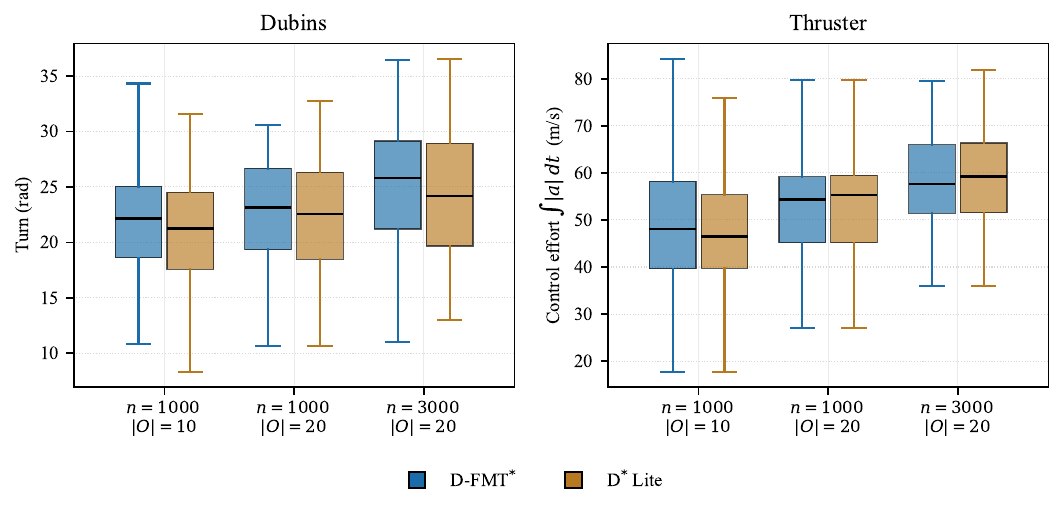}
	\caption{Fixed-graph benchmark. Domain-specific executed trajectory quality,
		reported as total heading change in Dubins and control effort
		$\int|a|\,dt$ in Thruster.}
	\label{fig:na_domain_quality}
\end{figure*}

\begin{figure*}[!t]
	\centering
	\includegraphics[width=\textwidth]{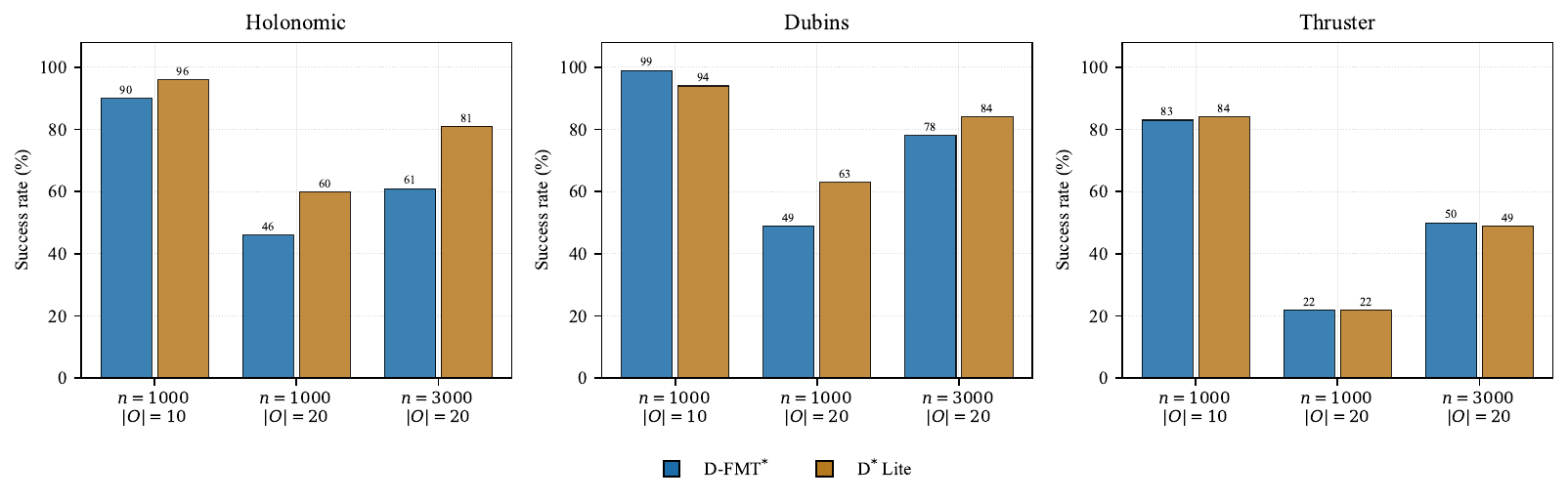}
	\caption{Fixed-graph benchmark. Success rate of D-FMT$^{*}$ and
		D$^{*}$~Lite over 100 seeds across graph size $n$ and obstacle count
		$|O|$.}
	\label{fig:na_success}
\end{figure*}

Figs.~\ref{fig:edge_checks_fixed} and~\ref{fig:na_repair_time}
show that D-FMT$^{*}$ uses far fewer collision checks and generally lower
median repair latency than D$^{*}$~Lite. The size of the advantage depends on
collision-check cost and neighborhood degree. Geometric $\mathbb{R}^{2}$ and
holonomic $\mathbb{R}^{2}{\times}T$ have cheap checks but high mean degree, so
D$^{*}$~Lite repeatedly processes large dynamic neighborhoods while
D-FMT$^{*}$ avoids most edge validation. The latency separation is more modest
in these spaces, and in the densest holonomic configuration
($n{=}3000$, $|O|{=}20$) the medians approach one another while the
D-FMT$^{*}$ tail grows as a high-degree cut drives a wider wavefront. Dubins
and Thruster have more expensive kinodynamic checks and lower degree, so
avoiding unnecessary validation gives D-FMT$^{*}$ its largest latency gains.

Executed quality remains close between the two planners
(Figs.~\ref{fig:na_path_length}--\ref{fig:na_domain_quality}). In Thruster,
the two are similar across configurations, with D-FMT$^{*}$ attaining lower
control effort at $n{=}3000$ and $|O|{=}20$. Elsewhere D-FMT$^{*}$ remains
close to D$^{*}$~Lite in executed path length, while Dubins heading change is
also similar. D-FMT$^{*}$ therefore achieves reductions of one to two orders
of magnitude in collision checks with only small differences in executed
quality, and the latency benefit is largest when individual checks are
expensive.

Where a robot is present, success requires maintaining a collision-free anchor
after each update (Fig.~\ref{fig:na_success}). Both planners perform well when
the PRM$^{*}$ is sufficiently connected and lose success when Thruster
coverage becomes sparse under $|O|{=}20$. On the harder Dubins and holonomic
configurations, D$^{*}$~Lite more often retains a feasible anchor. Conditional
on common success, D-FMT$^{*}$ remains the cheaper replanner.

D-FMT$^{*}$ therefore provides the fixed-graph repair foundation for
FMT\textsuperscript{X}. Its lazy parent validation substantially reduces
collision checking while preserving similar executed quality, with the main
limitation arising when the fixed graph offers too few feasible anchors.
Online densification in FMT\textsuperscript{X} addresses this limitation by
continuing to add reachable alternatives during execution.

\subsubsection{Anytime Replanning}
\label{subsec:anytime_replanning}

We compare FMT\textsuperscript{X}, RRT\textsuperscript{X}, and LLPT$^{*}$ as the
graph densifies online during execution.
Beyond the primary benchmark, two additional studies evaluate
time-to-goal reachability filtering and partial obstacle
observability under limited sensing.

\paragraph{Simulation I -- Default Anytime Baseline}
\label{subsubsec:anytime_sim1}

\begin{figure*}[!t]
	\centering
	\includegraphics[width=0.82\textwidth]{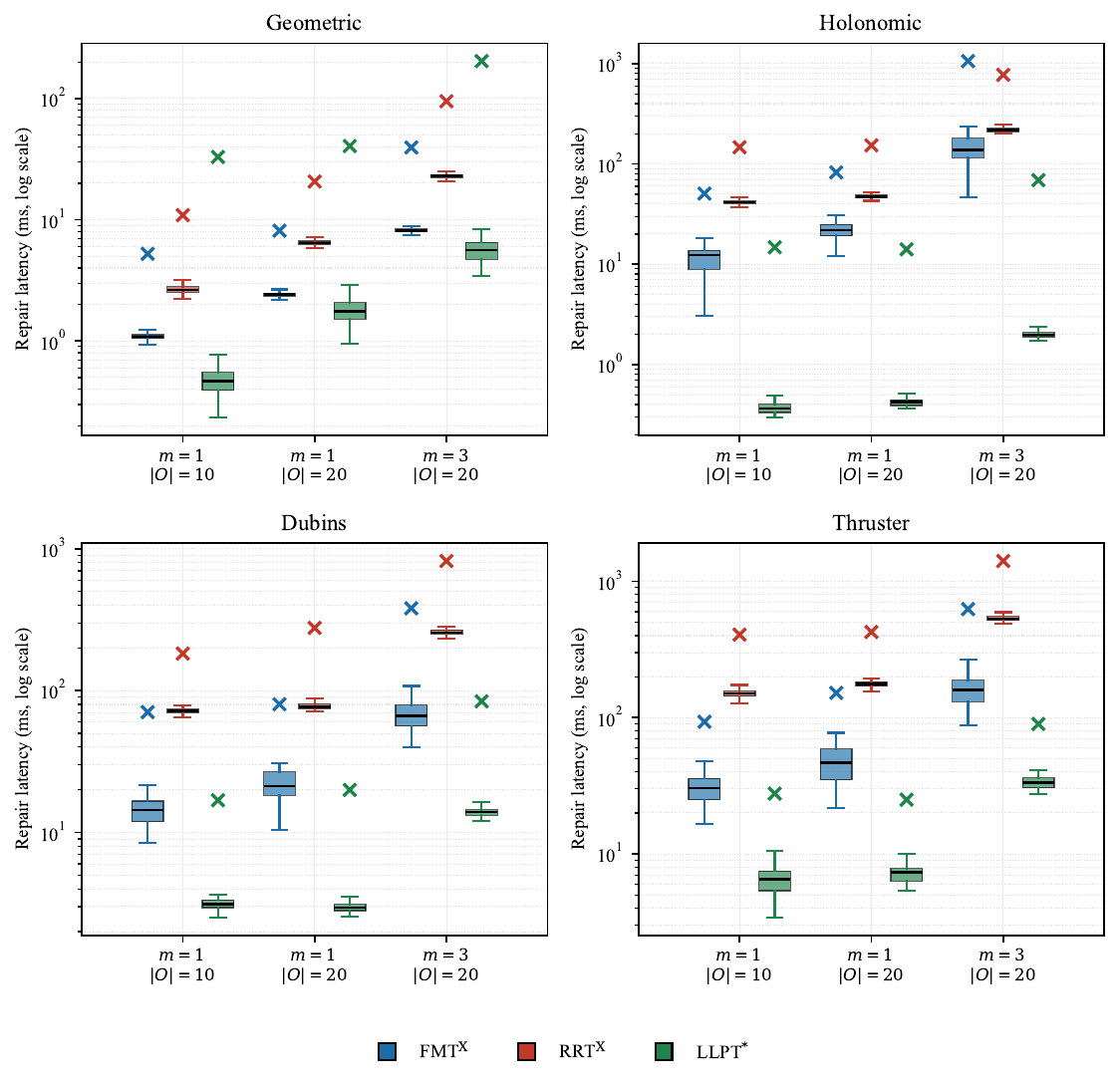}
	\caption{Simulation I. Repair latency of FMT\textsuperscript{X},
	RRT\textsuperscript{X}, and LLPT$^{*}$ across benchmark configurations.}
	\label{fig:anytime_sim1_repair}
\end{figure*}

The planners differ primarily in how they validate and repair the graph after
an obstacle event. LLPT$^{*}$ evaluates edges only when they appear on the
current optimistic candidate path, FMT\textsuperscript{X} detects affected tree
regions and repairs through cost-ordered parent candidates, and
RRT\textsuperscript{X} eagerly validates and restores local graph edges.
Together with differences in neighborhood degree and collision-check cost
across the state spaces, these policies produce the collision-check ordering
in Fig.~\ref{fig:edge_checks_sim1} and the repair latencies in
Fig.~\ref{fig:anytime_sim1_repair}. LLPT$^{*}$ generally has the lowest median
repair latency, while FMT\textsuperscript{X} requires substantially less repair
work than RRT\textsuperscript{X} in the more expensive kinodynamic settings.

As in the fixed-graph study, repair time reflects collision-check cost and
neighborhood degree. In geometric $\mathbb{R}^{2}$ and holonomic
$\mathbb{R}^{2}{\times}T$, checks are cheap but degree is high, so
RRT\textsuperscript{X} processes large neighborhoods while
FMT\textsuperscript{X} wavefronts can broaden after a cut, most visibly in the
densest holonomic setting. LLPT$^{*}$ keeps the lowest geometric median, but
has a heavier tail with larger maximum latencies. In Dubins and Thruster,
checks are more expensive and degree is lower, giving
FMT\textsuperscript{X} its clearest advantage over RRT\textsuperscript{X}.
Raising $m$ or $|O|$ increases absolute cost without changing these trends.
Initial planning time and terminal $|V|$ remain essentially matched, so the
differences are not caused by unequal sampling.

\begin{figure*}[!t]
	\centering
	\includegraphics[width=0.92\textwidth]{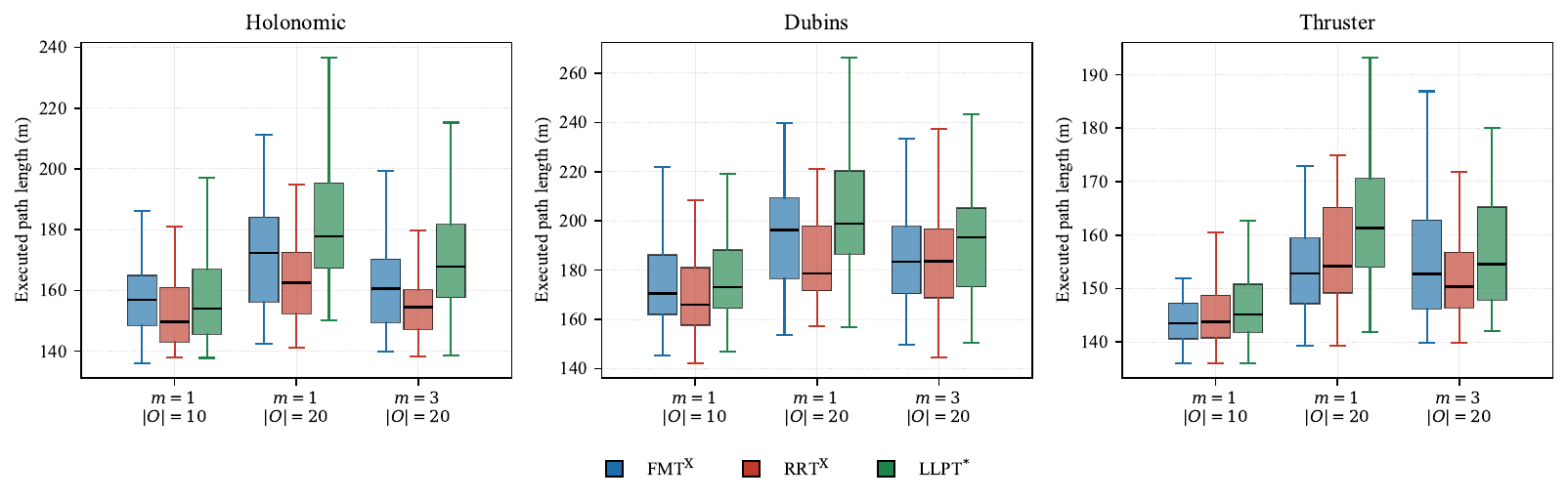}
	\caption{Simulation I. Executed path length of the anytime planners across
		the benchmark configurations.}
	\label{fig:anytime_sim1_path}
\end{figure*}

\begin{figure*}[!t]
	\centering
	\includegraphics[width=0.92\textwidth]{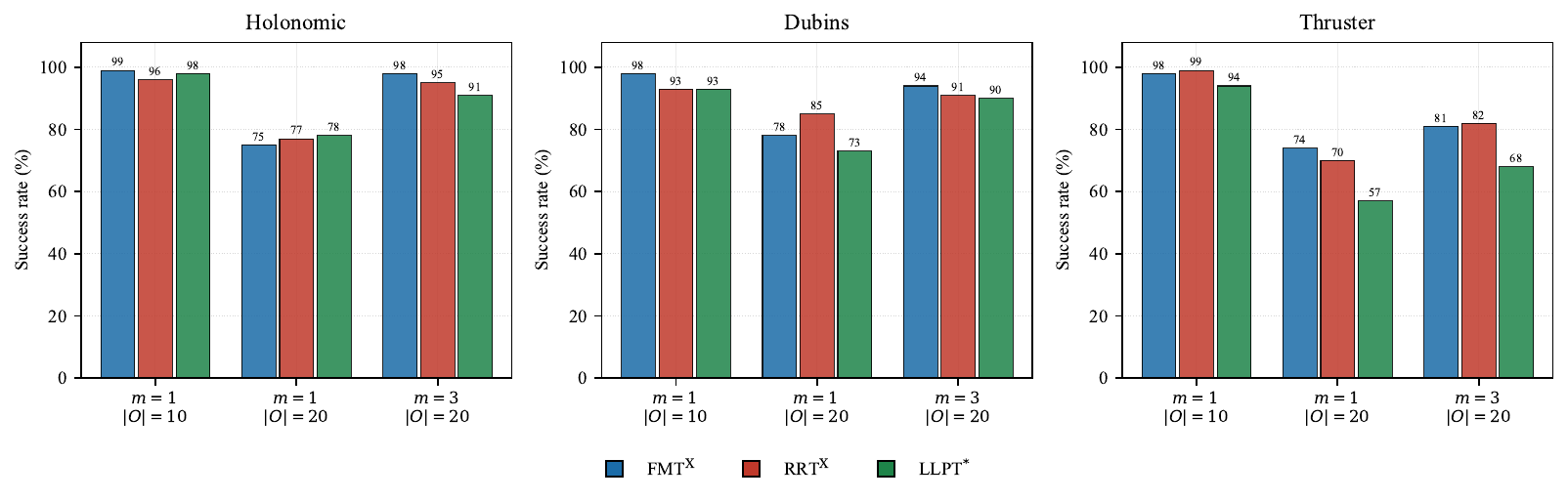}
	\caption{Simulation I. Success rate of the anytime planners over 100 seeds
		across the benchmark configurations.}
	\label{fig:anytime_sim1_success}
\end{figure*}

Low repair latency does not necessarily imply higher reliability
(Fig.~\ref{fig:anytime_sim1_success}). Under light dynamic load, all three
planners achieve high success. As obstacle density and kinodynamic difficulty
increase, LLPT$^{*}$ loses the most success, especially in Thruster, while
FMT\textsuperscript{X} and RRT\textsuperscript{X} remain more reliable.
Path-centric repair becomes less robust when updates repeatedly require
alternatives away from the current route.

Figure~\ref{fig:thruster_evolution} shows a representative Thruster run in
which the goal-rooted tree is repaired as the obstacles move.

\begin{figure*}[!t]
	\centering
	\includegraphics[width=0.88\textwidth]{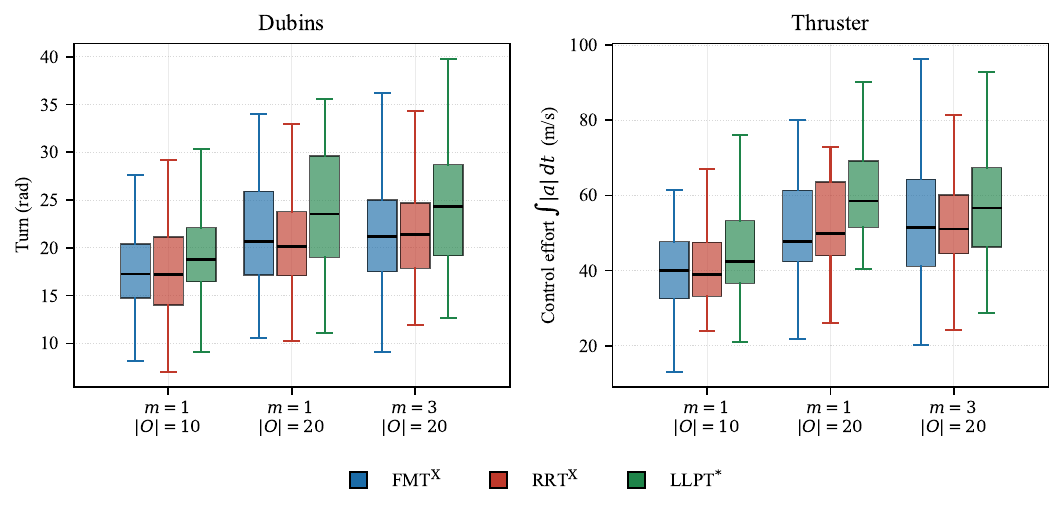}
	\caption{Simulation I. Domain-specific executed trajectory quality for the
		anytime planners, reported as total heading change in Dubins and control
		effort $\int |a|\,dt$ in Thruster.}
	\label{fig:anytime_sim1_domain}
\end{figure*}

Executed quality on jointly successful seeds reflects the amount of graph
consistency maintained by each repair policy
(Figs.~\ref{fig:anytime_sim1_path}--\ref{fig:anytime_sim1_domain}).
With a single root and fixed arrival budget, the main differences appear in
path length, heading change, and thruster effort rather than traversal time.
FMT\textsuperscript{X} and RRT\textsuperscript{X} generally retain better
executed quality than LLPT$^{*}$ when obstacle updates repeatedly invalidate
the active branch. Between the two graph-repairing planners,
FMT\textsuperscript{X} remains close to RRT\textsuperscript{X} in executed
quality while requiring far fewer collision checks, with the clearest
computational advantage in Dubins and Thruster.

Overall, LLPT$^{*}$ has the lowest typical repair cost,
RRT\textsuperscript{X} validates neighborhoods more extensively, and
FMT\textsuperscript{X} remains close to RRT\textsuperscript{X} in reliability
and executed quality on the harder Dubins and Thruster cases. The next two
studies examine reachability filtering and online obstacle revelation.

\FloatBarrier
\paragraph{Simulation II -- Time-to-Goal Reachability Filter}
\label{subsubsec:anytime_sim2}

We repeat the Thruster anytime setting with the time-to-goal
reachability filter of Section~\ref{sec:experimental_setup}.
Because filtered nodes lie outside the robot's remaining time budget, the
filter drops only provably irrelevant work and leaves feasible cost-to-go
unchanged on the live reachable subgraph.
We fix $m{=}1$ and $|O|{=}20$, and sweep
$T\in\{35,45,55,65\}\,\mathrm{s}$.
The $T{=}35\,\mathrm{s}$ cell matches Simulation~I so that only the filter
is varied.

\begin{figure*}[!t]
	\centering
	\begin{subfigure}[b]{0.48\textwidth}
		\centering
		\includegraphics[width=\textwidth]{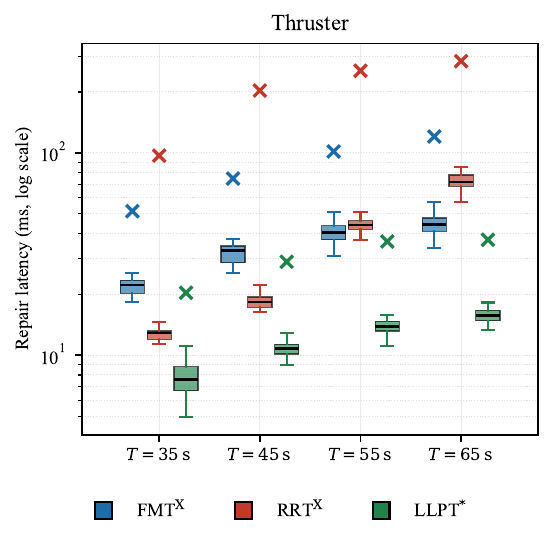}
		\caption{Repair latency.}
		\label{fig:anytime_sim2_repair}
	\end{subfigure}
	\hfill
	\begin{subfigure}[b]{0.48\textwidth}
		\centering
		\includegraphics[width=\textwidth]{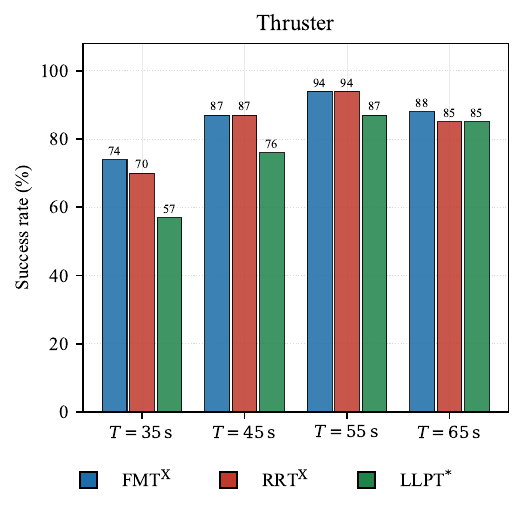}
		\caption{Success rate.}
		\label{fig:anytime_sim2_success}
	\end{subfigure}
	\vspace{0.5cm}
	\begin{subfigure}[b]{0.48\textwidth}
		\centering
		\includegraphics[width=\textwidth]{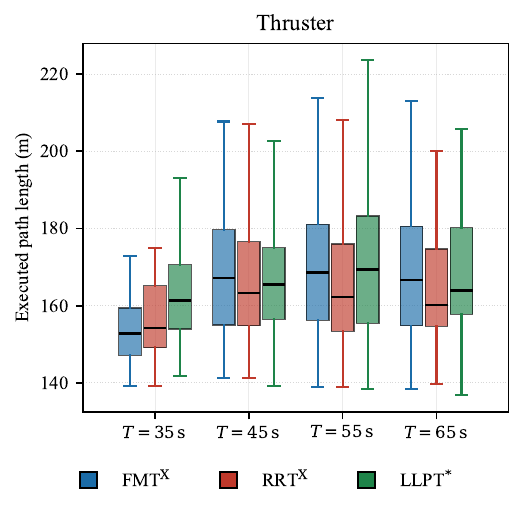}
		\caption{Executed path length.}
		\label{fig:anytime_sim2_path}
	\end{subfigure}
	\hfill
	\begin{subfigure}[b]{0.48\textwidth}
		\centering
		\includegraphics[width=\textwidth]{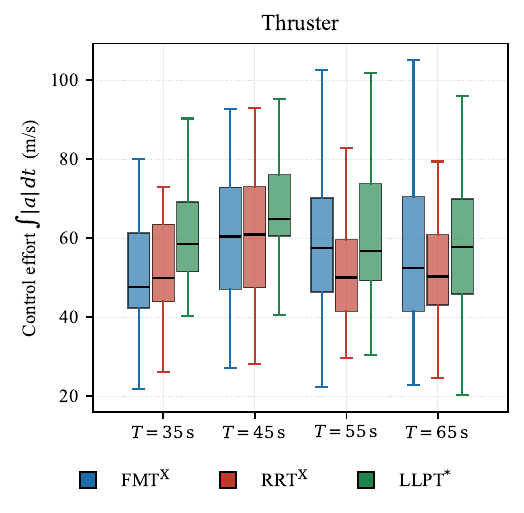}
		\caption{Domain-specific quality.}
		\label{fig:anytime_sim2_domain}
	\end{subfigure}
	\caption{Simulation II (Reachability Filter). Performance and quality metrics.}
	\label{fig:anytime_sim2_combined}
\end{figure*}

At matched $T{=}35\,\mathrm{s}$, success remains $74\%$, $70\%$, and $57\%$
for FMT\textsuperscript{X}, RRT\textsuperscript{X}, and LLPT$^{*}$,
respectively, and executed quality is unchanged from Simulation~I. Repair
latency and collision checks both decrease
(Figs.~\ref{fig:anytime_sim2_repair} and~\ref{fig:edge_checks_sim2}), with the
largest relative saving in RRT\textsuperscript{X}, a marked reduction in
FMT\textsuperscript{X}, and little change in LLPT$^{*}$.

The effect follows repair policy. RRT\textsuperscript{X} gains most because
the filter suppresses eager validation in regions the robot can no longer
reach. FMT\textsuperscript{X} also avoids work outside the live budget, but
its selective wavefront already touches fewer edges. LLPT$^{*}$ remains the
computational floor. In the tight-budget regime, LLPT$^{*}$ has the lowest
median latency, followed by RRT\textsuperscript{X} and then
FMT\textsuperscript{X}.

As $T$ grows, more of the graph remains within the live budget, so the filter
drops a smaller fraction of work, especially for RRT\textsuperscript{X}, whose
eager cost rises with live neighborhood volume. FMT\textsuperscript{X}
collision-check load levels off once the reachable graph stabilises, although
wall-clock repair still grows with densification and wavefront work.
LLPT$^{*}$ remains fastest, and its cost grows mainly with longer missions and
more path resolution. At larger $T$, FMT\textsuperscript{X} again becomes
faster than RRT\textsuperscript{X}. The separation between the two
graph-repairing planners therefore widens with the horizon rather than
remaining fixed. At the tighter budgets, RRT\textsuperscript{X} has lower median repair
latency than FMT\textsuperscript{X}, but substantially higher upper-tail and
maximum latency. As $T$ grows, FMT\textsuperscript{X} also becomes faster in
the median, and at the loosest budget it repairs appreciably faster than the
eager planner (Fig.~\ref{fig:anytime_sim2_repair}).

Success rises with manoeuvre room for all three planners and is highest in
the mid-budget cells. FMT\textsuperscript{X} and RRT\textsuperscript{X} stay
ahead of LLPT$^{*}$ throughout.

Overall, the reachability filter complements the underlying repair policy.
It helps eager neighborhood repair the most, changes path-centric laziness
the least, and leaves FMT\textsuperscript{X} with intermediate repair cost while maintaining
stable predictability and kinodynamic reliability.

\FloatBarrier

\paragraph{Simulation III -- Partial Observability}
\label{subsubsec:anytime_sim3}

The third simulation reveals static obstacles only within a fixed sensor
range (Fig.~\ref{fig:dubins_evolution}), so each discovery forces repair
against newly revealed geometry. We use Dubins with $m\in\{1,2\}$, the
multi-horizon goal set and reachability filter of
Sec.~\ref{sec:experimental_setup}, and a $100\,\mathrm{s}$ budget.
The primary signals are repair-latency tails and executed path length,
traversal time, and heading change on common-success seeds.
We do not compare planner success rates in this experiment. Unlike the
fixed-horizon fully observable benchmarks, the multi-horizon goal construction
allows different planners to reach the same spatial region at different
simulated times. Because samples are added online at a fixed rate per planning
slice, the graph density available when a difficult region is encountered can
therefore depend on the planner's preceding execution history.

This regime is harder for path-centric laziness than the fully observable
settings of Simulations~I--II. A planner that does not invalidate corridors
outside the current optimistic path can enter a region that is later revealed
to be a dead end. Recovery then requires repeated anchor search after several
optimistic candidates have already become invalid.

\begin{figure*}[!t]
	\centering
	\begin{subfigure}[b]{0.48\textwidth}
		\centering
		\includegraphics[width=\textwidth]{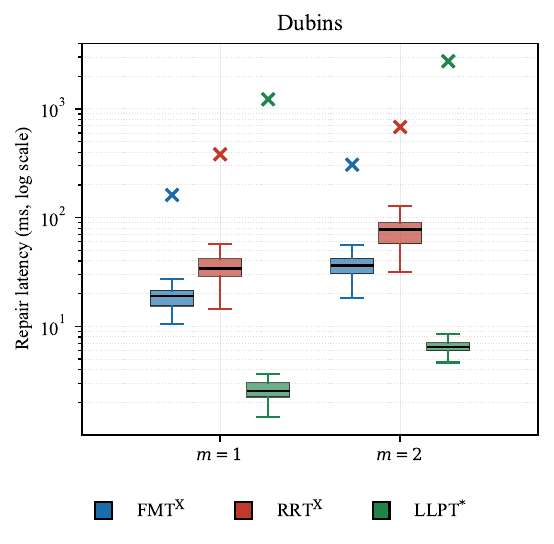}
		\caption{Repair latency.}
		\label{fig:anytime_sim3_repair}
	\end{subfigure}
	\hfill
	\vspace{0.5cm}
	\begin{subfigure}[b]{0.48\textwidth}
		\centering
		\includegraphics[width=\textwidth]{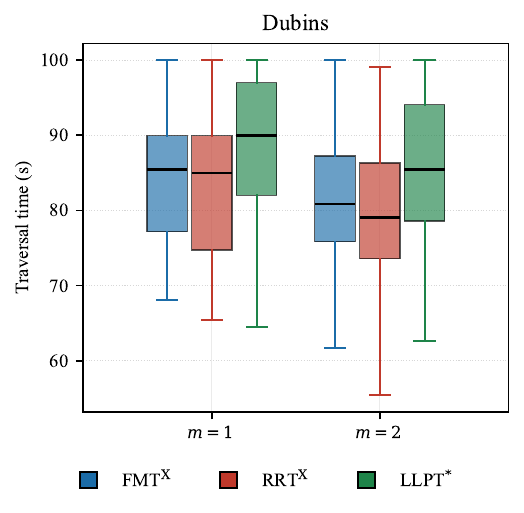}
		\caption{Traversal time.}
		\label{fig:anytime_sim3_exec_time}
	\end{subfigure}
	\hfill
	\begin{subfigure}[b]{0.48\textwidth}
		\centering
		\includegraphics[width=\textwidth]{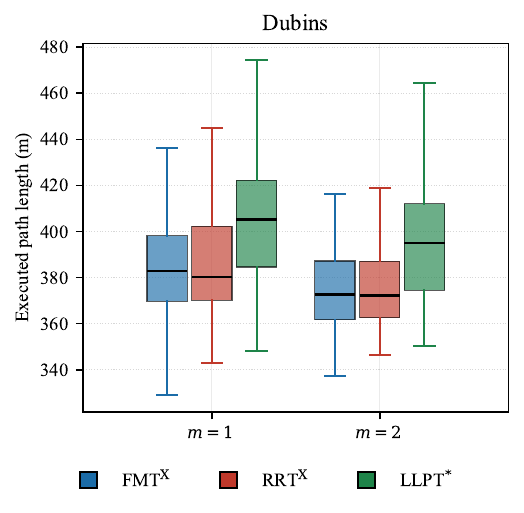}
		\caption{Executed path length.}
		\label{fig:anytime_sim3_path}
	\end{subfigure}
	\vspace{0.5cm}
	\begin{subfigure}[b]{0.48\textwidth}
		\centering
		\includegraphics[width=\textwidth]{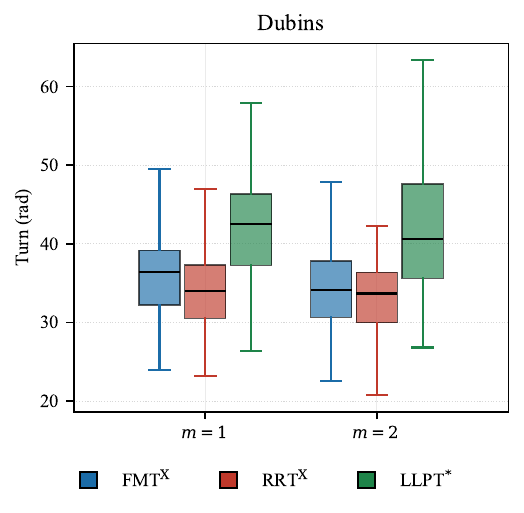}
		\caption{Domain-specific quality.}
		\label{fig:anytime_sim3_domain}
	\end{subfigure}
	\caption{Simulation III (Partial Observability). Performance and quality metrics.}
	\label{fig:anytime_sim3_combined}
\end{figure*}

Collision-check counts retain the familiar ordering
(Fig.~\ref{fig:edge_checks_sim3}), with LLPT$^{*}$ lowest,
RRT\textsuperscript{X} highest, and FMT\textsuperscript{X} in between.
Median repair latency follows the same broad ordering
(Fig.~\ref{fig:anytime_sim3_repair}), but the clearer distinction appears
in the tail. Path-centric repair defers invalidation until the active path is
resolved, so a newly revealed obstacle can be discovered only after the robot
has entered the affected corridor. Recovery then proceeds through failed
anchors and repeated lazy resolution, producing large repair-time spikes.
FMT\textsuperscript{X} and RRT\textsuperscript{X} instead invalidate the
affected subgraph at discovery time and repair alternatives before the robot
becomes trapped. Their median repair cost is higher, but the observed
high-percentile and maximum latencies are substantially lower.
FMT\textsuperscript{X} attains the lowest observed maximum repair latency of
the three planners, below RRT\textsuperscript{X} and roughly an order of
magnitude below LLPT$^{*}$.

Executed quality on jointly successful seeds shows the corresponding effect
(Figs.~\ref{fig:anytime_sim3_path}--\ref{fig:anytime_sim3_domain}).
LLPT$^{*}$ paths are longer, more turn-heavy, and finish later, consistent
with late dead-end discovery followed by larger geometric corrections rather
than slower tracking. FMT\textsuperscript{X} remains close to
RRT\textsuperscript{X} in path length and heading change. Longer LLPT$^{*}$
missions also accumulate more samples under per-slice densification, so
end-of-run graph size is not a clean comparator in this experiment.

Partial observability therefore separates the three validation strategies more
clearly than the fully observable benchmarks. LLPT$^{*}$ remains fastest on
typical slices but can pay heavily when previously unknown obstacles invalidate
the executed route. RRT\textsuperscript{X} repairs neighborhoods eagerly and
achieves the smoothest trajectories at the highest median repair cost.
FMT\textsuperscript{X} again requires less validation than RRT\textsuperscript{X} while reducing the observed
latency tail relative to LLPT$^{*}$ and remaining close to RRT\textsuperscript{X} in trajectory
quality. In sensor-limited operation, when collision checks are performed
can therefore matter as much as how many are performed.

\FloatBarrier

\subsection{Trade-off Summary}
\label{sec:tradeoffs}

Across the fixed-graph and anytime benchmarks, the main distinction is not
whether repair is localized, but how collision checking is scheduled within
the affected region. D-FMT$^{*}$ and FMT\textsuperscript{X} check selected
parent connections during cost-ordered wavefront repair, D$^{*}$~Lite and
RRT\textsuperscript{X} evaluate affected neighborhood connections more eagerly,
and LLPT$^{*}$ postpones validation until edges appear on an optimistic path.

On a shared PRM$^{*}$, D-FMT$^{*}$ consistently performs far fewer edge checks
than D$^{*}$~Lite and has lower median repair time when individual checks are
expensive. The separation is largest for Dubins and Thruster and more moderate
for geometric and holonomic graphs, where checks are cheap but neighborhood
degree is high. On the densest holonomic cells, D-FMT$^{*}$ wavefronts can
broaden the latency tail, while D$^{*}$~Lite remains more predictable.
Among the anytime planners, LLPT$^{*}$ performs the least validation,
RRT\textsuperscript{X} performs the most through eager neighborhood repair,
and FMT\textsuperscript{X} repairs affected subgraphs with selective parent
validation.

Median repair time and tail behavior do not always agree. LLPT$^{*}$ is
typically the fastest, but in the Dubins partial-observability experiment its
deferred validation produces large repair-time spikes after previously unknown
obstacles invalidate the active route. FMT\textsuperscript{X} and
RRT\textsuperscript{X} instead invalidate affected structure when the obstacle
is revealed, substantially reducing the observed tail.

Overall, FMT\textsuperscript{X} uses substantially less validation than eager
neighborhood repair while remaining close to RRT\textsuperscript{X} in
reliability and executed trajectory quality.

\section{Conclusion}
\label{sec:conclusion}

This work introduced D-FMT$^{*}$ and FMT\textsuperscript{X}, extending the
lazy, cost-ordered wavefront of FMT$^{*}$ to dynamic replanning. Obstacle
changes trigger localized repair of the affected subgraph, with rewiring under
a cost-improvement condition and collision checking only on selected parent
connections. D-FMT$^{*}$ applies this mechanism to a fixed sampled graph,
while FMT\textsuperscript{X} additionally densifies the graph online. The theoretical analysis establishes asymptotic optimality under the
assumptions of Section~\ref{sec:runtime_analysis}, with expected
$O(n\log n+nT_{\mathrm{coll}})$ repair work and $O(n)$ collision checks.

On shared PRM$^{*}$ graphs, D-FMT$^{*}$ uses substantially fewer collision
checks than D$^{*}$~Lite, with the clearest repair-time gains when edge
evaluation is expensive. The anytime experiments reveal a trade-off among
LLPT$^{*}$, FMT\textsuperscript{X}, and RRT\textsuperscript{X}.
LLPT$^{*}$ performs the fewest collision checks,
RRT\textsuperscript{X} repairs more eagerly at higher cost, and
FMT\textsuperscript{X} lies between them while remaining competitive in
trajectory quality and reliability. Path-centric laziness can require repeated
resolution when deferred collision checks invalidate successive candidate
shortest paths before a collision-free path is found. Under partial
observability, late obstacle discovery can amplify this effect and produce
large repair-latency spikes.

Overall, the results show that effective lazy replanning depends not only on
deferring collision checks, but also on how deferred validation is integrated
with graph repair. FMT\textsuperscript{X} combines selective parent checking with subgraph-level wavefront repair,
requiring less validation than eager neighborhood repair while avoiding the
strictly path-centric validation strategy of LLPT$^{*}$.

\FloatBarrier

\appendix
\renewcommand{\thesection}{Appendix~\Alph{section}}
\setcounter{equation}{0}
\renewcommand{\theequation}{\Alph{section}.\arabic{equation}}
\section{Proofs}
\label{app:proofs}

\begin{proof}[Proof of Lemma~\ref{lem:termination}]
	Fix the snapshot. Labels are reset only between intervals, so the argument is
	local to one drain. Every accepted relaxation writes
	$\mathrm{lmc}(x)\gets\mathrm{lmc}(y)+d_\pi(x,y)$ for some $y$ and, by
	\textup{(P3)}, only when the written value is strictly below the current
	$\mathrm{lmc}(x)$. Such a write never closes a cycle, because if the path realizing
	$\mathrm{lmc}(y)$ already passed through $x$, then positive edge costs give
	$\mathrm{lmc}(y)>\mathrm{lmc}(x)$, hence
	$\mathrm{lmc}(y)+d_\pi(x,y)>\mathrm{lmc}(x)$ and the guard fails. Every finite
	label is therefore the cost of a simple path in $\mathcal{G}$ terminating at
	$x_{\mathrm{goal}}$, and on a finite vertex set there are finitely many such
	costs. By \textup{(P3)} each accepted relaxation strictly decreases one
	coordinate of the label vector and increases none, so the vector descends
	through a finite set of values and only finitely many relaxations are
	accepted. Queue insertions follow either the finitely many update-time seeds
	or an accepted relaxation, so the queue empties.
\end{proof}

\begin{proof}[Proof of Lemma~\ref{lem:consistency}]
	Induction over intervals.

	\textbf{First interval.} From the initial labeling of \textup{(H2)}, every
	vertex other than $x_{\mathrm{goal}}$ carries $\mathrm{lmc}=\infty$. A vertex
	acquires a finite label only through a write, which by \textup{(P3)} queues
	it, and the drain is full, so every vertex with a finite label at the end of
	the first interval was extracted during it. Case~1 below therefore covers
	every pair, and the case for an unextracted $y$ is vacuous. Note that this
	argument needs neither Lemma~\ref{lem:cost-equivalence} nor an obstacle-free
	environment.

	\textbf{Later intervals.} Assume~\eqref{eq:consistency} at the end of the
	previous interval, and suppose for contradiction that it fails at the end of
	the current one for some pair $(x,y)$ with $y\in N^{+}(x)$ and $(x,y)$ free in
	the current snapshot, the pair $(x,y)$ carrying no unresolved suboptimal connection in the current interval or
	in any preceding one. That is,
	\begin{equation}
		\label{eq:contra}
		\ell(x)\;>\;\ell(y)+d_\pi(x,y).
	\end{equation}

	Note that~\eqref{eq:contra} forces $\ell(y)<\infty$, so every pair examined
	below has a finite candidate parent, and orphans with no finite neighbor never
	arise here.

	\textbf{Case 1.} Suppose $y$ is extracted during the drain.
	Consider the last extraction of $y$ within the current drain. By \textup{(P3)} that extraction sets
	$g(y)=\mathrm{lmc}(y)$, and no later write occurs, so $g(y)=\ell(y)$. Labels
	are monotone, so the label of $x$ at that instant is at least $\ell(x)$, and
	by~\eqref{eq:contra} it exceeds $g(y)+d_\pi(x,y)$. Since $y\in N^{+}(x)$ and
	$x\in N^{-}(y)$ record the same retained connection at its two endpoints, the
	reverse-neighborhood sweep of Algorithm~\ref{alg:Expand} required by
	\textup{(P3)} reaches $x$ and its guard fires. By
	\textup{(P3)} the parent search then installs a label at most
	$g(y)+d_\pi(x,y)$, since $y$ itself is an open candidate at that instant.
	Hence $\ell(x)\le\ell(y)+d_\pi(x,y)$, contradicting~\eqref{eq:contra}, unless
	the argmin failed its collision check, which is the stated exception.

	\textbf{Case 2.} Suppose $y$ is never extracted during the drain.
	By \textup{(P3)} with $\epsilon=0$ no write to $\mathrm{lmc}(y)$ occurred,
	since under that setting any write queues $y$ and the full drain would extract
	it. Hence
	$\ell(y)=\ell_0(y)$, and $y$ is not orphaned.

	Suppose first that $x$ is orphaned by the update, or that $(x,y)$ was blocked
	in the previous snapshot. These two triggers correspond to the two obstacle
	operations, and they are what \textup{(P2)} is stated to cover. An insertion
	only shrinks free space, so by \textup{(P1)} it can act on a surviving finite
	label only by orphaning, while a removal only enlarges free space, so it can
	act only by freeing a previously blocked edge. Property \textup{(P2)} applies
	directly to the pair $(x,y)$, since the edge is free in the current snapshot,
	$y\in N^{+}(x)$, and $\ell_0(y)<\infty$ whenever the right-hand side
	of~\eqref{eq:contra} is finite, and places
	$y\in S\subseteq V_{\mathrm{open}}$. The drain is full, so $y$ is extracted,
	contradicting the case assumption. Only $y$ need be seeded, since $x$ is
	relaxed passively when $y$ is extracted and its sweep of $N^{-}(y)$ reaches $x$.

	Otherwise $x$ was present in the previous interval, is not orphaned, and
	$(x,y)$ was free in the previous snapshot.
	Then $\ell_0(x)$ and $\ell_0(y)$ are the previous interval's terminal labels,
	so the induction hypothesis applies to the pair and gives
	$\ell_0(x)\le\ell_0(y)+d_\pi(x,y)$. Monotonicity \textup{(P3)} yields
	$\ell(x)\le\ell_0(x)\le\ell(y)+d_\pi(x,y)$, again
	contradicting~\eqref{eq:contra}.
\end{proof}

\begin{proof}[Proof of Lemma~\ref{lem:no-defer}]
	Let $(x,y)=(x_m,x_{m-1})$ be a consecutive chain pair and let $k$ be the
	interval under analysis. Suppose, for contradiction, that a parent search at
	$x$ triggered by $y$ was resolved by a failed collision check at some earlier
	interval, and let $j<k$ be the last such interval. The failure occurred on a
	candidate edge $(x,y_{\min})$ selected while $y$ was the extracted candidate,
	so $d_\pi(x,y_{\min})\le d_\pi(x,y)\le r_n$ by \textup{(C1)} and the
	queue-ordering bound, and the check failed because that edge met some obstacle
	$o$ present in snapshot $j$. Hence $\mathrm{dist}(x,o)\le r_n$.

	Since the failed connection has length at most $r_n$ and is free in snapshot
	$k$ by \textup{(C2)}, there is some update $i$ with $j<i\le k$ at which that
	stored connection becomes free. Property \textup{(P2)} then reseeds its
	finite-$\mathrm{lmc}$ neighborhood.

	If $y$ already has finite $\mathrm{lmc}$ at that update, it is seeded and
	extracted by the full drain. Otherwise the first later write that makes
	$\mathrm{lmc}(y)$ finite queues $y$ by \textup{(P3)}, so $y$ is likewise
	extracted before the end of the drain in which that write occurs. Since $\ell(y)<\infty$ in interval
	$k$, such an extraction occurs no later than $k$.

	At that extraction, either the guard at $x$ does not fire, in which case the
	then-current labels satisfy $\mathrm{lmc}(x)\le g(y)+d_\pi(x,y)$ and the
	inherited pairwise inconsistency is repaired, or it fires. In the latter case
	the parent search is triggered by $y$. If this extraction occurs before
	interval $k$, another failed check would contradict the choice of $j$ as the
	last such failed search before $k$. If it occurs in interval $k$, the
	queue-ordering bound again gives
	$d_\pi(x,y_{\min})\le d_\pi(x,y)\le r_n$, and \textup{(C2)} rules out a
	collision failure. Thus the search succeeds, and Case~1 of
	Lemma~\ref{lem:consistency} repairs the pairwise inconsistency. Hence the
	pair cannot carry an unresolved inherited exception in interval $k$.
\end{proof}

\begin{proof}[Proof of Lemma~\ref{lem:path-bound}]
	Work on $G_n$. Under \textup{(H1)} the setting is geometric, so $d_\pi$ is the
	Euclidean distance and the neighborhoods are Euclidean $r_n$-balls. By
	\textup{(C1)} each chain edge $(x_m,x_{m+1})$ is free with
	$\|x_m-x_{m+1}\|\le r_n$, so $x_m\in N^{+}(x_{m+1})$. By
	Lemma~\ref{lem:no-defer} the exception of Lemma~\ref{lem:consistency} fires for
	no consecutive chain pair, so the lemma applies to $(x_{m+1},x_m)$, giving
	\begin{equation*}
		\ell(x_{m+1})\;\le\;\ell(x_m)+\|x_m-x_{m+1}\| .
	\end{equation*}
	Starting from $\ell(x_1)=0$ by \textup{(H3)} and telescoping,
	$\ell(x_m)\le C_m$ for every $m$, so
	$c_n=\ell(x_{M_n})\le C_{M_n}\le L_n$ by \textup{(C3)}. The bound therefore
	fails only on $G_n^c$, so $P(c_n>L_n)\le P(G_n^c)\to0$
	by~\cite[Lem.~4.3--4.4]{janson2015fast}.
\end{proof}

\section{Anytime Edge-Check}
\label{app:anytime-checks}

Across all anytime simulations, LLPT$^{*}$ performs the fewest collision
checks, RRT\textsuperscript{X} the most, and FMT\textsuperscript{X} lies
between them.

\begin{figure}[!ht]
	\centering
	\includegraphics[width=\textwidth]{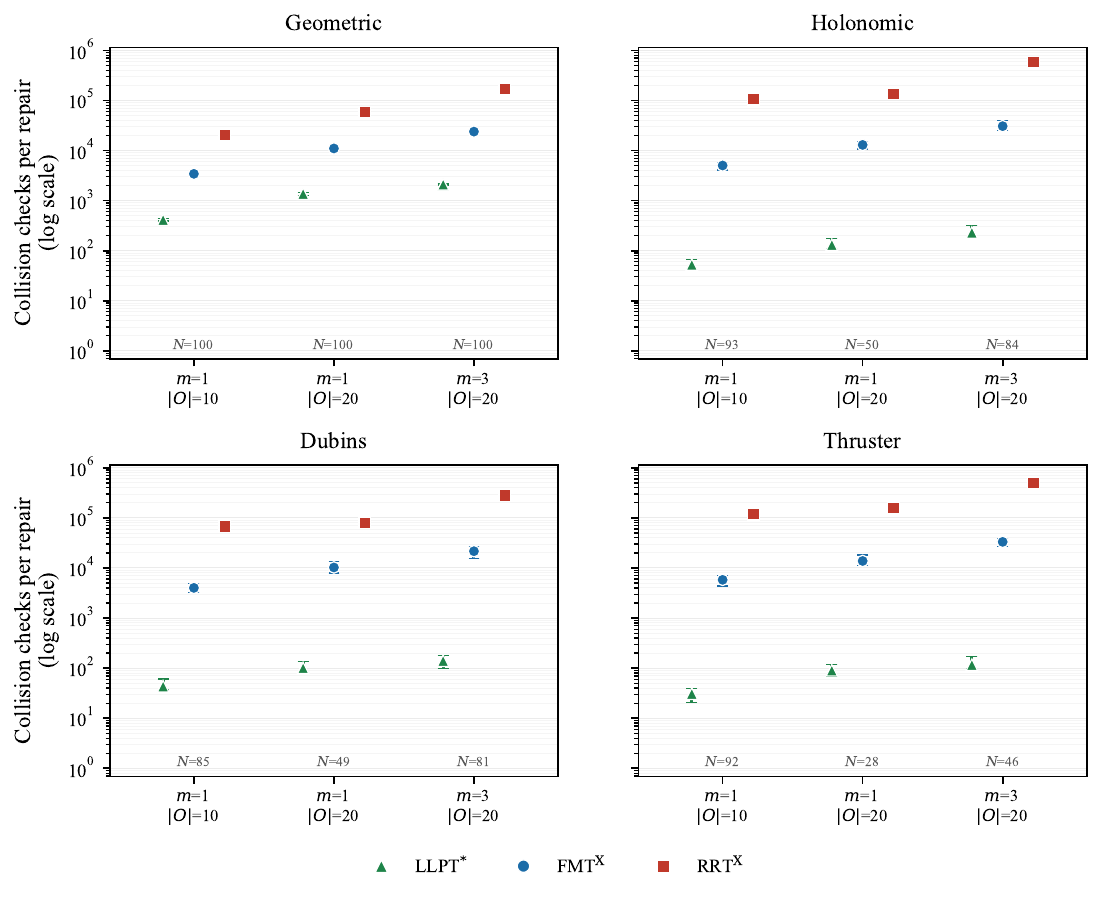}
	\caption{Simulation~I. Collision checks per repair on a logarithmic scale. Points show the
		median of the per-seed mean collision checks per repair over paired
		common-success seeds, and error bars denote Q1--Q3. $N$
		denotes the number of paired common-success seeds used for that
		configuration.}
	\label{fig:edge_checks_sim1}
\end{figure}

\begin{figure}[!ht]
	\centering
	\begin{subfigure}[b]{0.545\textwidth}
		\centering
		\includegraphics[width=\textwidth]{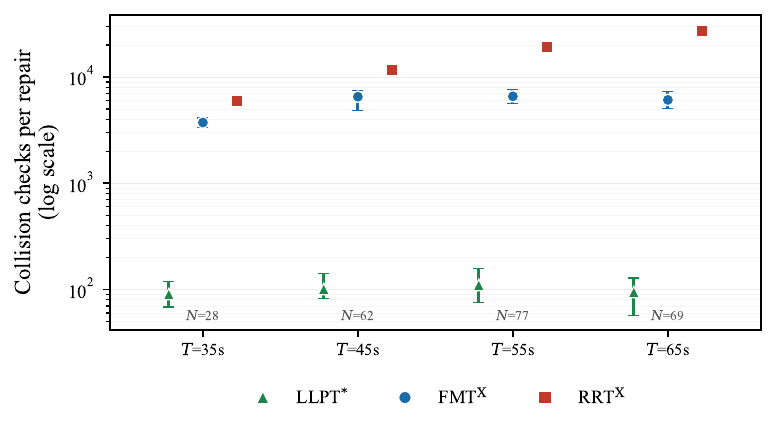}
		\caption{Simulation~II.}
		\label{fig:edge_checks_sim2}
	\end{subfigure}
	\hfill
	\begin{subfigure}[b]{0.425\textwidth}
		\centering
		\includegraphics[width=\textwidth]{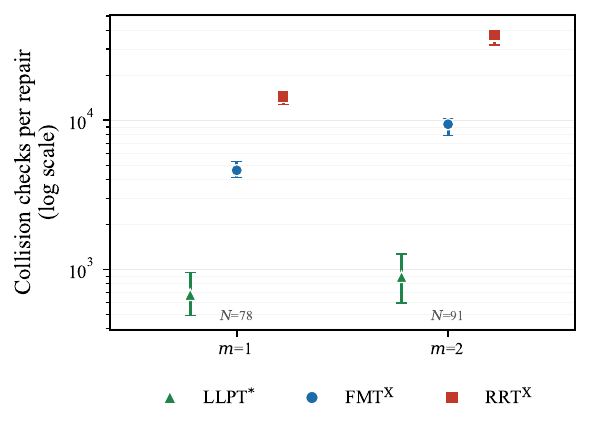}
		\caption{Simulation~III.}
		\label{fig:edge_checks_sim3}
	\end{subfigure}
	\caption{Simulations~II and~III. Collision checks per repair on a logarithmic scale. Points show the
		median of the per-seed mean collision checks per repair over paired
		common-success seeds, and error bars denote Q1--Q3. $N$
		denotes the number of paired common-success seeds used for that
		configuration.}
	\label{fig:edge_checks_sim23}
\end{figure}

\FloatBarrier

\end{document}